# Learning One-hidden-layer ReLU Networks via Gradient Descent

Xiao Zhang*† and Yaodong Yu*‡ and Lingxiao Wang*§ and Quanquan Gu¶

## Abstract

We study the problem of learning one-hidden-layer neural networks with Rectified Linear Unit (ReLU) activation function, where the inputs are sampled from standard Gaussian distribution and the outputs are generated from a noisy teacher network. We analyze the performance of gradient descent for training such kind of neural networks based on empirical risk minimization, and provide algorithm-dependent guarantees. In particular, we prove that tensor initialization followed by gradient descent can converge to the ground-truth parameters at a linear rate up to some statistical error. To the best of our knowledge, this is the first work characterizing the recovery guarantee for practical learning of one-hidden-layer ReLU networks with multiple neurons. Numerical experiments verify our theoretical findings.

## 1 Introduction

Deep neural networks have achieved lots of breakthroughs in the field of artificial intelligence, such as speech recognition (Hinton et al., 2012), image processing (Krizhevsky et al., 2012), statistical machine translation (Bahdanau et al., 2014), and Go games (Silver et al., 2016). The empirical success of neural networks stimulates numerous theoretical studies in this field. For example, in order to explain the superiority of neural networks, a series of work (Hornik, 1991; Barron, 1993; Daniely et al., 2016; Cohen et al., 2016) investigated the expressive power of neural networks. It has been proved that given appropriate weights, neural networks with nonlinear activation function, such as sigmoid, can approximate any continuous function.

In practice, (stochastic) gradient descent remains one of the most widely-used approaches for deep learning. However, due to the nonconvexity and nonsmoothness of the loss function landscape, existing theory in optimization cannot explain why gradient-based methods can effectively learn neural networks. To bridge this gap, a line of research (Tian, 2017; Li and Yuan, 2017; Du et al., 2017a,b) studied (stochastic) gradient descent for learning shallow neural networks from a theoretical perspective. More specifically, by assuming an underlying teacher network, they established recovery guarantees for applying gradient-based learning algorithms to the population loss function (a.k.a., expected risk function). Another line of research (Zhong et al., 2017; Soltanolkotabi, 2017; Soltanolkotabi et al., 2017; Fu et al., 2018) investigated using (stochastic) gradient descent to minimize the empirical loss function of shallow neural networks, and provided theoretical guarantees on sample complexity, i.e., number of samples required for recovery.

---

*Equal Contribution

†Department of Computer Science, University of Virginia, Charlottesville, VA 22904, USA; e-mail: xz7bc@virginia.edu

‡Department of Computer Science, University of Virginia, Charlottesville, VA 22904, USA; e-mail: yy8ms@virginia.edu

§Department of Computer Science, University of Virginia, Charlottesville, VA 22904, USA; e-mail: lw4wr@virginia.edu

¶Department of Computer Science, University of Virginia, Charlottesville, VA 22904, USA; e-mail: qg5w@virginia.edu

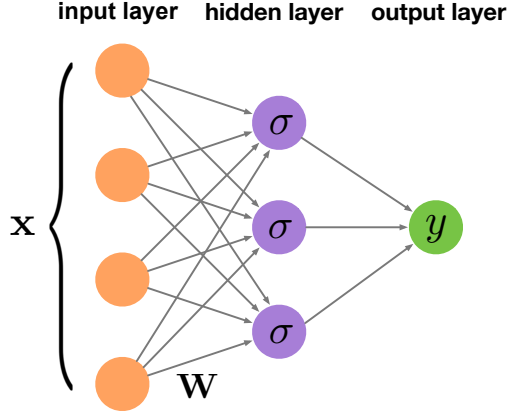

Figure 1: Illustration of one-hidden-layer ReLU-based teacher network (1.1).

Our work follows the second line of research, where we directly study the empirical risk minimization of one-hidden-layer ReLU networks, and characterize the recovery guarantee using gradient descent. More specifically, we assume the inputs $\{\mathbf{x}_i\}_{i=1}^N \subseteq \mathbb{R}^d$ follow standard multivariate Gaussian distribution, and the outputs $\{y_i\}_{i=1}^N \subseteq \mathbb{R}$ are generated from the following one-hidden-layer ReLU-based teacher network (see Figure 1 for graphical illustration)

$$y_i = \sum_{j=1}^K \sigma(\mathbf{w}_j^{*\top}\mathbf{x}_i) + \epsilon_i, \quad \text{for any } i \in [N]. \tag{1.1}$$

Here, $\mathbf{w}_j^* \in \mathbb{R}^d$ denotes the weight parameter with respect to the $j$-th neuron, $\sigma(x) = \max\{x, 0\}$ denotes the ReLU activation function, and $\{\epsilon_i\}_{i=1}^N$ are i.i.d. zero mean sub-Gaussian random noises[1] with sub-Gaussian norm $\nu > 0$. Our goal is to recover the unknown parameter matrix $\mathbf{W}^* = [\mathbf{w}_1^*, \ldots, \mathbf{w}_K^*] \in \mathbb{R}^{d \times K}$ based on the observed $N$ examples $(\mathbf{x}_i, y_i)_{i=1}^N$. The recovery problem can be equivalently formulated as the following empirical risk minimization problem using square loss

$$\min_{\mathbf{W} \in \mathbb{R}^{d \times K}} \widehat{\mathcal{L}}_N(\mathbf{W}) = \frac{1}{2N} \sum_{i=1}^N \bigg(\sum_{j=1}^K \sigma(\mathbf{w}_j^\top \mathbf{x}_i) - y_i\bigg)^2, \tag{1.2}$$

where $\mathbf{W} = [\mathbf{w}_1, \ldots, \mathbf{w}_K]$. In this paper, we show that with good starting point and sample complexity linear in $d$, using gradient descent to solve (1.2) is guaranteed to converge to $\mathbf{W}^*$ at a linear rate. To the best of our knowledge, this is the first result of its kind to prove the theoretical guarantee for learning one-hidden-layer ReLU networks with multiple neurons based on the empirical loss function. We believe our analysis on one-hidden-layer ReLU networks can shed light on the understanding of gradient-based methods for learning deeper neural networks. The main contributions of this work are summarized as follows:

- We consider the empirical risk minimization problem (1.2) for learning one-hidden-layer ReLU networks. Compared with existing studies (Tian, 2017; Li and Yuan, 2017; Du et al., 2017a,b) that consider the ideal population risk minimization, our analysis is more aligned with the practice of deep learning that is based on the empirical loss function.

- We analyze the performance of gradient descent based algorithm for minimizing the empirical loss function. We demonstrate that, provided an appropriate initial solution, gradient descent can linearly converge to the ground-truth parameters of the underlying teacher network (1.1) up to some statistical error. In particular, the statistical error term depends on the sample size $N$, the input dimension $d$, the number of neurons in the hidden layer $K$, as well as the magnitude of the noise distribution $\nu$. In addition, we show that the sample complexity for recovery required by our algorithm is linear in $d$ up to a logarithmic factor.

- We provide a uniform convergence bound on the gradient of the empirical loss function (1.2). More specifically, we characterize the difference between the gradient of the empirical loss function and the gradient of the population loss function, when the parameters are close to the ground-truth parameters. This result enables us to establish the linear convergence guarantee of gradient descent method without using resampling (i.e., sample splitting) trick adopted in Zhong et al. (2017).

---

[1] The formal definitions of sub-Gaussian random variable and sub-Gaussian norm can be found in Section 4.

The remainder of this paper is organized as follows: In Section 2, we discuss the most related literature to our work. We introduce the problem setup and our proposed algorithm in Section 3. We present the main theoretical results and their proof in Sections 4 and 5 respectively. In Section 6, we conduct experiments to verify our theory. Finally, we conclude our paper and discuss some future work in Section 7.

**Notation.** We use $[d]$ to denote the set $\{1, 2, \ldots, d\}$. For any $d$-dimensional vector $\mathbf{x} = [x_1, \ldots, x_d]^\top$, let $\|\mathbf{x}\|_2 = (\sum_{i=1}^d |x_i|^2)^{1/2}$ be its $\ell_2$ norm. For any matrix $\mathbf{A} = [A_{ij}]$, denote the spectral norm and Frobenius norm of $\mathbf{A}$ by $\|\mathbf{A}\|_2$ and $\|\mathbf{A}\|_F$, respectively. Let $\sigma_{\max}(\mathbf{A})$, $\sigma_{\min}(\mathbf{A})$ be the largest singular value and smallest singular value of $\mathbf{A}$, respectively. Given any two sequences $\{a_n\}$ and $\{b_n\}$, we write $a_n = O(b_n)$ if there exists a constant $0 < C < +\infty$ such that $a_n \leq C b_n$, and we use $\widetilde{O}(\cdot)$ to hide the logarithmic factors. We use $\mathbb{1}\{\mathcal{E}\}$ to denote the indicator function such that $\mathbb{1}\{\mathcal{E}\} = 1$ if the event $\mathcal{E}$ is true, otherwise $\mathbb{1}\{\mathcal{E}\} = 0$. For two matrices $\mathbf{A}, \mathbf{B}$, we say $\mathbf{A} \succeq \mathbf{B}$ if $\mathbf{A} - \mathbf{B}$ is positive semidefinite. We use $\mathcal{B}_r(\mathbf{B}) = \{\mathbf{A} \in \mathbb{R}^{d \times K} : \|\mathbf{A} - \mathbf{B}\|_F \leq r\}$ to denote the Frobenius norm ball centering at $\mathbf{B}$ with radius $r$.

## 2 Related Work

To better understand the extraordinary performance of neural networks on different tasks, a line of research (Hornik, 1991; Montufar et al., 2014; Cohen et al., 2016; Telgarsky, 2016; Raghu et al., 2016; Poole et al., 2016; Arora et al., 2016; Daniely et al., 2016; Pan and Srikumar, 2016; Zhang et al., 2016; Lu et al., 2017) has studied on the expressive power of neural networks. In particular, Hornik (1991) showed that, with sufficient number of neurons, shallow networks can approximate any continuous function. Cohen et al. (2016); Telgarsky (2016) proved that a shallow network requires exponential size to realize functions that can be implemented by a deep network of polynomial size. Raghu et al. (2016); Poole et al. (2016); Arora et al. (2016) characterized the exponential dependence on the depth of the network based on different measures of expressivity. Recently, Zhang et al. (2016) empirically demonstrated that neural networks can actually memorize the training samples but still generalize well. Daniely et al. (2016); Lu et al. (2017) showed how the depth and width can affect the expressive power of neural networks.

However, the expressive power of neural networks can only partially explain the empirical success of deep learning. From a theoretical perspective, it is well-known that learning neural networks in general settings is hard in the worst case (Blum and Rivest, 1989; Auer et al., 1996; Livni et al., 2014; Shamir, 2016; Shalev-Shwartz et al., 2017a,b; Zhang et al., 2017; Ge et al., 2017). Nevertheless, a vast literature (Kalai and Sastry, 2009; Kakade et al., 2011; Sedghi and Anandkumar, 2014; Janzamin et al., 2015; Zhang et al., 2015; Goel et al., 2016; Arora et al., 2016) developed ad hoc algorithms that can learn neural networks with provable guarantees. However, none of these algorithms is gradient-based method, which is the most widely-used optimization algorithm for deep learning in practice.

Recently, a series of work (Tian, 2017; Brutzkus and Globerson, 2017; Li and Yuan, 2017; Du et al., 2017a,b) studied the recovery guarantee of gradient-based methods for learning shallow neural networks based on population loss function (i.e., expected risk function). More specifically, Tian (2017) proved that for one-layer one-neuron ReLU networks (i.e., ReLU unit), randomly initialized gradient descent on the population loss function can recover the groundtruth parameters of the teacher network. In a concurrent work, Brutzkus and Globerson (2017) considered the problem of learning a convolution filter and showed that gradient descent enables exact recovery of the true parameters, provided the filters are non-overlapping. Later on, Li and Yuan (2017) studied one-hidden-layer residual networks, and proved that stochastic gradient descent can recover the underlying true parameters in polynomial number of iterations. Du et al. (2017a) studied the convergence of gradient-based methods for learning a convolutional filter. They showed that under certain conditions, performing (stochastic) gradient descent on the expected risk function can recover the underlying true parameters in polynomial time. Du et al. (2017b) further studied the problem of learning the one-hidden-layer ReLU based convolutional neural network in the no-overlap patch setting. More specifically, they established the convergence guarantee of gradient descent with respect to the expected risk function when the input follows Gaussian distribution. Nevertheless, all these studies are based on the population loss function.

In practice, training neural networks is based on the empirical loss function. To the best of our knowledge, only several recent studies (Zhong et al., 2017; Soltanolkotabi, 2017; Soltanolkotabi et al., 2017; Fu et al., 2018) analyzed gradient based methods for training neural networks using empirical risk minimization. More specifically, under condition that the activation function is smooth, Zhong et al. (2017); Soltanolkotabi et al. (2017); Fu et al. (2018) established a locally linear convergence rate for gradient descent with suitable initialization scheme. However, none of their analyses are applicable to ReLU networks since ReLU activation function is nonsmooth[2]. Soltanolkotabi (2017) analyzed the projected gradient descent on the empirical loss function for one-neuron ReLU networks (i.e., ReLU unit). Yet the analysis requires a projection step to ensure convergence, while the constraint set of the projection depends on the unknown ground-truth weight vector, which makes their algorithm less practical. Our work also follows this line of research, where we investigate the theoretical performance of gradient descent for learning one-hidden-layer ReLU

---

[2]While many activation functions including ReLU are discussed in Zhong et al. (2017), their locally linear convergence result for gradient descent is not applicable to ReLU activation function.

networks with multiple neurons.

Inspired by the success of first-order optimization algorithms (Ge et al., 2015; Jin et al., 2017) for solving nonconvex optimization problems efficiently, some recent work (Choromanska et al., 2015; Safran and Shamir, 2016; Mei et al., 2016; Kawaguchi, 2016; Hardt and Ma, 2016; Soltanolkotabi et al., 2017; Soudry and Carmon, 2016; Xie et al., 2017; Nguyen and Hein, 2017; Ge et al., 2017; Safran and Shamir, 2017; Yun et al., 2017; Du and Lee, 2018) attempted to understand neural networks by characterizing their optimization landscape. Choromanska et al. (2015) studied the loss surface of a special random neural network. Safran and Shamir (2016) analyzed the geometric structure of the over-parameterized neural networks. Mei et al. (2016) studied the landscape of the empirical loss of the one-layer neural network given that the third derivative of the activation function is bounded. Kawaguchi (2016) showed that there is no spurious local minimum for linear deep networks. Hardt and Ma (2016) proved that linear residual networks have no spurious local optimum. Soudry and Carmon (2016); Xie et al. (2017); Nguyen and Hein (2017); Yun et al. (2017) also showed that there is no spurious local minimum for some other neural networks under stringent assumptions. Soltanolkotabi et al. (2017) studied the global optimality of the over-parameterized network with quadratic activation functions. On the other hand, Ge et al. (2017); Safran and Shamir (2017) showed that ReLU neural networks with multiple neurons using square loss actually have spurious local minima. To address this issue, Ge et al. (2017) proposed to modify the objective function of ReLU networks, and showed that the modified objective function has no spurious local minimum, thus perturbed gradient descent can learn the groundtruth parameters. Compared with Ge et al. (2017), we directly analyze the objective function of ReLU networks based on square loss using gradient descent without modifying the loss function, but needing a special initialization.

## 3 Problem Setup and Algorithm

In this section, we present the problem formulation along with a gradient descent-based algorithm for learning one-hidden-layer ReLU networks. Recall that our goal is to recover the unknown parameter matrix $\mathbf{W}^*$ based on the empirical loss function in (1.2). For the ease of later analysis, we define the corresponding population loss function as follows

$$\mathcal{L}(\mathbf{W}) = \frac{1}{2}\mathbb{E}_{\mathbf{X} \sim \mathcal{D}_X}\bigg(\sum_{j=1}^K \sigma(\mathbf{w}_j^\top \mathbf{X}) - \sum_{j=1}^K \sigma(\mathbf{w}_j^{*\top} \mathbf{X})\bigg)^2, \tag{3.1}$$

where $\mathcal{D}_X = N(\mathbf{0}, \mathbf{I})$ denotes the standard multivariate Gaussian distribution. In addition, let $\sigma_1 \geq \sigma_2 \geq \ldots \geq \sigma_K > 0$ be the sorted singular values of $\mathbf{W}^*$, and $\kappa = \sigma_1/\sigma_K$ be the condition number of $\mathbf{W}^*$, and $\lambda = (\Pi_{j=1}^K \sigma_j)/\sigma_K^K$.

In this work, we focus on minimizing the empirical loss function in (1.2) instead of the population loss function in (3.1), because in practice one can only get access to the training data $\{(\mathbf{x}_i, y_i)\}_{i=1}^N$. Witnessing the empirical success of the widely-used gradient-based methods for training neural networks, one natural question is whether gradient descent can recover $\mathbf{W}^*$ based on the empirical loss function in (1.2). In later analysis, we will show that the answer to the above question is affirmative. The gradient descent algorithm for solving the nonconvex and nonsmooth optimization problem (1.2) is demonstrated in Algorithm 1.

It is worth noting that the gradient descent algorithm shown in Algorithm 1 does not require any resampling (a.k.a., sample splitting) procedure (Jain et al., 2013) compared with the gradient

**Algorithm 1** Gradient Descent

**Input:** empirical loss function $\widehat{\mathcal{L}}_N$; step size $\eta$; iteration number $T$; initial estimator $\mathbf{W}^0$.
    **for** $t = 1, 2, 3, \ldots, T$ **do**
        $\mathbf{W}^t = \mathbf{W}^{t-1} - \eta \nabla \widehat{\mathcal{L}}_N(\mathbf{W}^{t-1})$
    **end for**
**Output:** $\mathbf{W}^T$.

---

descent algorithm analyzed in Zhong et al. (2017). More specifically, the gradient descent algorithm in Zhong et al. (2017) requires a fresh subset of the whole training sample at each iteration in order to establish the convergence guarantee. In sharp contrast, Algorithm 1 analyzed in this paper does not need resampling. The reason is that we are able to establish a uniform convergence bound between the gradient of the empirical loss function and the gradient of the population loss function, as will be illustrated in the next section. Furthermore, we lay out the explicit form of the derivative of $\widehat{\mathcal{L}}_N(\mathbf{W})$ with respect to $\mathbf{w}_k$ as follows

$$\left[\nabla \widehat{\mathcal{L}}_N(\mathbf{W})\right]_k = \sum_{j=1}^{K} \left( \widehat{\mathbf{\Sigma}}(\mathbf{w}_j, \mathbf{w}_k) \mathbf{w}_j - \widehat{\mathbf{\Sigma}}(\mathbf{w}_j^*, \mathbf{w}_k) \mathbf{w}_j^* \right) - \frac{1}{N} \sum_{i=1}^{N} \epsilon_i \mathbf{x}_i \cdot \mathbb{1}\{\mathbf{w}_k^\top \mathbf{x}_i \geq 0\}, \quad (3.2)$$

where $\widehat{\mathbf{\Sigma}}(\mathbf{w}_j, \mathbf{w}_k)$ and $\widehat{\mathbf{\Sigma}}(\mathbf{w}_j^*, \mathbf{w}_k)$ are defined as

$$\begin{aligned}
\widehat{\mathbf{\Sigma}}(\mathbf{w}_j, \mathbf{w}_k) &= \frac{1}{N} \sum_{i=1}^{N} \left[ \mathbf{x}_i \mathbf{x}_i^\top \cdot \mathbb{1}\{\mathbf{w}_j^\top \mathbf{x}_i \geq 0, \mathbf{w}_k^\top \mathbf{x}_i \geq 0\} \right], \\
\widehat{\mathbf{\Sigma}}(\mathbf{w}_j^*, \mathbf{w}_k) &= \frac{1}{N} \sum_{i=1}^{N} \left[ \mathbf{x}_i \mathbf{x}_i^\top \cdot \mathbb{1}\{\mathbf{w}_j^{*\top} \mathbf{x}_i \geq 0, \mathbf{w}_k^\top \mathbf{x}_i \geq 0\} \right].
\end{aligned} \quad (3.3)$$

## 4 Main Theory

Before presenting our main theoretical results, we first lay out the definitions of sub-Gaussian random variable and sub-Gaussian norm.

**Definition 4.1.** (sub-Gaussian random variable) We say $X$ is a sub-Gaussian random variable with sub-Gaussian norm $K > 0$, if $(\mathbb{E}|X|^p)^{1/p} \leq K\sqrt{p}$ for all $p \geq 1$. In addition, the sub-Gaussian norm of $X$, denoted $\|X\|_{\psi_2}$, is defined as $\|X\|_{\psi_2} = \sup_{p \geq 1} p^{-1/2}(\mathbb{E}|X|^p)^{1/p}$.

The following theorem shows that as long as the initial estimator $\mathbf{W}^0$ falls in a small neighbourhood of $\mathbf{W}^*$, gradient descent algorithm in Algorithm 1 is guaranteed to converge to $\mathbf{W}^*$ with a linear rate of convergence.

**Theorem 4.2.** Assume the inputs $\{\mathbf{x}_i\}_{i=1}^{N}$ are sampled from standard Gaussian distribution, and the outputs $\{y_i\}_{i=1}^{N}$ are generated from the teacher network (1.1). Suppose the initial estimator $\mathbf{W}^0$ satisfies $\|\mathbf{W}^0 - \mathbf{W}^*\|_F \leq c\sigma_K/(\lambda \kappa^3 K^2)$, where $c > 0$ is a small enough absolute constant. Then there exist absolute constants $c_1, c_2, c_3, c_4$ and $c_5$ such that provide the sample size satisfies

$$N \geq \frac{c_1 \lambda^4 \kappa^{10} K^9 d}{\sigma_K^2} \log\left(\frac{\lambda \kappa K d}{\sigma_K}\right) \cdot (\|\mathbf{W}^*\|_F^2 + \nu^2),$$

the output of Algorithm 1 with step size $\eta \leq 1/(c_2\kappa K^2)$ satisfies

$$\|\mathbf{W}^T - \mathbf{W}^*\|_F^2 \leq \left(1 - \frac{c_3\eta}{\lambda\kappa^2}\right)^T \|\mathbf{W}^0 - \mathbf{W}^*\|_F^2 + \frac{c_4\lambda^2\kappa^4 K^5 d\log N}{N} \cdot (\|\mathbf{W}^*\|_F^2 + \nu^2) \quad (4.1)$$

with probability at least $1 - c_5/d^{10}$.

**Remark 4.3.** Theorem 4.2 suggests that provided that the initial solution $\mathbf{W}_0$ is sufficiently close to $\mathbf{W}^*$, the output of Algorithm 1 exhibits a linear convergence towards $\mathbf{W}^*$, up to some statistical error. More specifically, the estimation error is bounded by two terms (see the right hand side of (4.1)): the first term is the optimization error, and the second term represents the statistical error. The statistical error depends on the sample size $N$, the input dimension $d$, the number of neurons in the hidden layer $K$ and some other problem-specific parameters. In addition, due to the existence of statistical error, we are only able to achieve at best $\varepsilon = c\lambda^2\kappa^4 K^5(\|\mathbf{W}^*\|_F^2 + \nu^2) \cdot (d\log N/N)$ estimation error, where $c$ is an absolute constant. Because of the linear convergence rate, it is sufficient to perform $T = O(\lambda\kappa^3 K^2 \cdot \log(1/\varepsilon))$ number of iterations in Algorithm 1 to make sure the optimization error is less than $\varepsilon$. Putting these pieces together gives the overall sample complexity of Algorithm 1 to achieve $\varepsilon$-estimation error:

$$O\bigg(\text{poly}(\lambda, \kappa, K, \sigma_K, \nu, \|\mathbf{W}^*\|_F) \cdot d\log\big(\lambda\kappa Kd/(\sigma_K)\big) \log(1/\varepsilon)\bigg).$$

Apparently, it is in the order of $\widetilde{O}(\text{poly}(K) \cdot d)$ if we treat other problem-specific parameters as constants. Correspondingly, the statistical error is in the order of $\widetilde{O}(\text{poly}(K) \cdot d/N)$.

The remaining question is how to find a good initial solution $\mathbf{W}^0$ for Algorithm 1, which satisfies the assumption of Theorem 4.2. The following lemma, proved in Zhong et al. (2017), suggests that we can obtain such desired initial estimator using tensor initialization.

**Lemma 4.4.** (Zhong et al., 2017) Consider the empirical risk minimization in (1.2), if the sample size $N \geq \epsilon^{-2} \cdot d \cdot \text{poly}(\kappa, K, \log d)$, with probability at least $1 - d^{-10}$, the output $\mathbf{W}^0 \in \mathbb{R}^{d \times K}$ of the tensor initialization satisfies

$$\|\mathbf{W}^0 - \mathbf{W}^*\|_F \leq \epsilon \cdot \text{poly}(\kappa, K)\|\mathbf{W}^*\|_F.$$

**Remark 4.5.** According to Lemma 4.4, if we set the approximation error $\epsilon$ such that

$$\epsilon \leq \frac{c_1\sigma_K}{\lambda\kappa^3 K^2 \text{poly}(\kappa, K)\|\mathbf{W}^*\|_F},$$

where $c_1$ is an absolute constant, the initial estimator $\mathbf{W}^0$ satisfies the assumption of Theorem 4.2. The corresponding sample complexity requirement for tensor initialization is in the order of $O(\text{poly}(\lambda, \kappa, K, \|\mathbf{W}^*\|_F, \log d) \cdot d)$. Therefore, combining Lemma 4.4 with Theorem 4.2, we conclude that tensor initialization followed by gradient descent can learn one-hidden-layer ReLU networks with linear convergence rate and overall sample complexity $\widetilde{O}(\text{poly}(K) \cdot d)$.

# 5 Proof of the Main Theory

In this section, we lay out the proof of our main result. To prove Theorem 4.2, we need to make use of the following lemmas. Lemmas 5.1 and 5.2 characterize the local strong convexity and smoothness of the population loss function around $\mathbf{W}^*$. Lemma 5.3 provides a uniform convergence bound on the difference between the gradient of the empirical loss function and the gradient of the population loss function in terms of Frobenius norm.

**Lemma 5.1.** For any $\mathbf{W} \in \mathbb{R}^{d \times K}$ such that $\|\mathbf{W} - \mathbf{W}^*\|_F \leq c\sigma_K/(\lambda \kappa^3 K^2)$, the Hessian of the population loss function $\mathcal{L}(\mathbf{W})$ satisfies

$$\nabla^2 \mathcal{L}(\mathbf{W}) \succeq \mu \mathbf{I},$$

where $\mu = c/(\lambda \kappa^2)$ and $c > 0$ is an absolute constant.

**Lemma 5.2.** (Safran and Shamir, 2017) The gradient of the population loss function $\nabla \mathcal{L}(\cdot)$ is $L$-Lipschitz within the region $\Omega = \{\mathbf{W} \in \mathbb{R}^{d \times K} \mid \|\mathbf{W} - \mathbf{W}^*\|_F \leq \sigma_K/2\}$, i.e., for any $\mathbf{W}_1, \mathbf{W}_2 \in \Omega$

$$\|\nabla \mathcal{L}(\mathbf{W}_1) - \nabla \mathcal{L}(\mathbf{W}_2)\|_F \leq L\|\mathbf{W}_1 - \mathbf{W}_2\|_F,$$

where $L = c\kappa K^2$, and $c > 0$ is an absolute constant.

**Lemma 5.3.** Consider the empirical loss function $\widehat{\mathcal{L}}_N(\mathbf{W})$ in (1.2). For all $\mathbf{W} \in \mathbb{R}^{d \times K}$ such that $\|\mathbf{W} - \mathbf{W}^*\|_F \leq c\sigma_K/(\lambda \kappa^3 K^2)$, where $c$ is an absolute constant, there exists absolute constants $c_1, c_2$ such that with probability at least $1 - c_1/d^{10}$, we have

$$\|\nabla \widehat{\mathcal{L}}_N(\mathbf{W}) - \nabla \mathcal{L}(\mathbf{W})\|_F \leq c_2 K^{5/2} \sqrt{\frac{d \log N}{N}} (\|\mathbf{W}^*\|_F + \nu),$$

where $\nu$ is the sub-Gaussian norm of the additive noise in the teacher network.

Now we are ready to prove the main theorem.

*Proof of Theorem 4.2.* We prove it by induction. We make the following inductive hypothesis

$$\|\mathbf{W}^t - \mathbf{W}^*\|_F \leq c\sigma_K/(\lambda \kappa^3 K^2). \tag{5.1}$$

Note that based on the assumption of Theorem 4.2, the initial estimator $\mathbf{W}^0$ satisfies (5.1), thus it remains to prove the inductive step. In other words, we need to show that $\mathbf{W}^{t+1}$ satisfies (5.1), provided that (5.1) holds for $\mathbf{W}^t$. Consider the gradient-based Algorithm 1 at the $(t+1)$-th iteration, we have

$$\mathbf{W}^{t+1} = \mathbf{W}^t - \eta \nabla \widehat{\mathcal{L}}_N(\mathbf{W}^t),$$

which implies that

$$\|\mathbf{W}^{t+1} - \mathbf{W}^*\|_F^2 = \|\mathbf{W}^t - \mathbf{W}^*\|_F^2 - 2\eta \langle \nabla \widehat{\mathcal{L}}_N(\mathbf{W}^t), \mathbf{W}^t - \mathbf{W}^* \rangle + \eta^2 \|\nabla \widehat{\mathcal{L}}_N(\mathbf{W}^t)\|_F^2$$
$$\leq \|\mathbf{W}^t - \mathbf{W}^*\|_F^2 \underbrace{- 2\eta \langle \nabla \mathcal{L}(\mathbf{W}^t), \mathbf{W}^t - \mathbf{W}^* \rangle + 2\eta^2 \|\nabla \mathcal{L}(\mathbf{W}^t)\|_F^2}_{I_1}$$
$$\underbrace{- 2\eta \langle \nabla \widehat{\mathcal{L}}_N(\mathbf{W}^t) - \nabla \mathcal{L}(\mathbf{W}^t), \mathbf{W}^t - \mathbf{W}^* \rangle + 2\eta^2 \|\nabla \widehat{\mathcal{L}}_N(\mathbf{W}^t) - \nabla \mathcal{L}(\mathbf{W}^t)\|_F^2}_{I_2},$$

where the inequality follows from $(a-b)^2 \le 2a^2 + 2b^2$.

In the following discussions, we are going to bound the terms $I_1$ and $I_2$, respectively. Consider the first term $I_1$. Note that according to the population loss function (3.1), we have $\nabla \mathcal{L}(\mathbf{W}^*) = \mathbf{0}$. Thus, we have

$$\text{vec}(\nabla \mathcal{L}(\mathbf{W}^t)) = \text{vec}(\nabla \mathcal{L}(\mathbf{W}^t) - \nabla \mathcal{L}(\mathbf{W}^*))$$
$$= \int_0^1 \nabla^2 \mathcal{L}(\mathbf{W}^* + \theta(\mathbf{W}^t - \mathbf{W}^*))d\theta \cdot \text{vec}(\mathbf{W}^t - \mathbf{W}^*) = \mathbf{H}_t \text{vec}(\mathbf{W}^t - \mathbf{W}^*),$$

where $\mathbf{H}_t = \int_0^1 \nabla^2 \mathcal{L}(\mathbf{W}^* + \theta(\mathbf{W}^t - \mathbf{W}^*))d\theta \in \mathbb{R}^{dK \times dK}$. Note that by the inductive assumption, we have $\|\mathbf{W}^t - \mathbf{W}^*\|_F \le c\sigma_K/(\lambda \kappa^3 K^2) \le \sigma_K/2$. Thus, according to Lemma 5.2, we obtain the upper bound of $I_1$

$$I_1 = -2\eta \cdot \text{vec}(\mathbf{W}^t - \mathbf{W}^*)^\top \mathbf{H}_t \text{vec}(\mathbf{W}^t - \mathbf{W}^*) + 2\eta^2 \cdot \text{vec}(\mathbf{W}^t - \mathbf{W}^*)^\top \mathbf{H}_t^\top \mathbf{H}_t \text{vec}(\mathbf{W}^t - \mathbf{W}^*)$$
$$\le (-2\eta + 2L\eta^2) \cdot \text{vec}(\mathbf{W}^t - \mathbf{W}^*)^\top \mathbf{H}_t \text{vec}(\mathbf{W}^t - \mathbf{W}^*),$$

where the inequality follows from Lemma 5.2 and $L = c_1 \kappa K^2$ is the Lipschitz parameter of $\nabla \mathcal{L}(\cdot)$. On the other hand, as for the term $I_2$, we have with probability at least $1 - c_3/d^{10}$ that

$$I_2 \le 2(\eta\beta + \eta^2) \cdot \|\nabla \widehat{\mathcal{L}}_N(\mathbf{W}^t) - \nabla \mathcal{L}(\mathbf{W}^t)\|_F^2 + \frac{2\eta}{\beta} \cdot \|\mathbf{W}^t - \mathbf{W}^*\|_F^2$$
$$\le 2(\eta\beta + \eta^2) \cdot \frac{c_2^2 K^5 d \log N}{N}(\|\mathbf{W}^*\|_F + \nu)^2 + \frac{2\eta}{\beta} \cdot \|\mathbf{W}^t - \mathbf{W}^*\|_F^2,$$

where the first inequality holds due to the Young's inequality, $\beta > 0$ is a constant that will be specified later, and the second inequality follows from Lemma 5.3. Set $\eta \le 1/(2L)$ and $\beta = 4/\mu$. Under condition that $\|\mathbf{W}^t - \mathbf{W}^*\|_F \le c\sigma_K/(\lambda\kappa^3 K^2)$, we obtain

$$\|\mathbf{W}^{t+1} - \mathbf{W}^*\|_F^2 \le \left(1 + \frac{2\eta}{\beta}\right) \cdot \|\mathbf{W}^t - \mathbf{W}^*\|_F^2 + (-2\eta + 2L\eta^2) \cdot \text{vec}(\mathbf{W}^t - \mathbf{W}^*)^\top \mathbf{H}_t \text{vec}(\mathbf{W}^t - \mathbf{W}^*)$$
$$+ 2(\eta\beta + \eta^2) \cdot \frac{c_2^2 K^5 d \log N}{N}(\|\mathbf{W}^*\|_F + \nu)^2$$
$$\le \left(1 + \frac{2\eta}{\beta} + \mu(-2\eta + 2L\eta^2)\right)\|\mathbf{W}^t - \mathbf{W}^*\|_F^2 + \frac{9c_2^2 \eta K^5 d \log N}{\mu N}(\|\mathbf{W}^*\|_F + \nu)^2$$
$$\le \left(1 - \frac{\mu\eta}{2}\right) \cdot \|\mathbf{W}^t - \mathbf{W}^*\|_F^2 + \frac{18c_2^2 \eta K^5 d \log N}{\mu N} \cdot (\|\mathbf{W}^*\|_F^2 + \nu^2) \quad (5.2)$$

holds with probability at least $1 - c_3/d^{10}$, where $\mu$ is the lower bound of the smallest singular value of $\nabla^2 \mathcal{L}(\mathbf{W})$ as in Lemma 5.1, the second inequality follows from Lemma 5.1. Hence, as long as the sample size satisfies

$$N \ge \frac{c_4 \lambda^2 \kappa^6 K^9 d}{\sigma_K^2 \mu^2} \log\left(\frac{\lambda \kappa K d}{\sigma_K \mu}\right) \cdot (\|\mathbf{W}^*\|_F^2 + \nu^2),$$

we have $\mathbf{W}^{t+1}$ satisfies (5.1). Thus, we proved the inductive hypothesis.

Finally, we conclude that with probability at least $1 - c_3/d^{10}$

$$\|\mathbf{W}^T - \mathbf{W}^*\|_F^2 \le \left(1 - \frac{\mu\eta}{2}\right)^T \|\mathbf{W}^0 - \mathbf{W}^*\|_F^2 + \frac{c_5 K^5 d \log N}{\mu^2 N} \cdot (\|\mathbf{W}^*\|_F^2 + \nu^2),$$

where we utilize the fact that $L = c_1 \kappa K^2$ as in Lemma 5.2, which completes the proof. $\square$

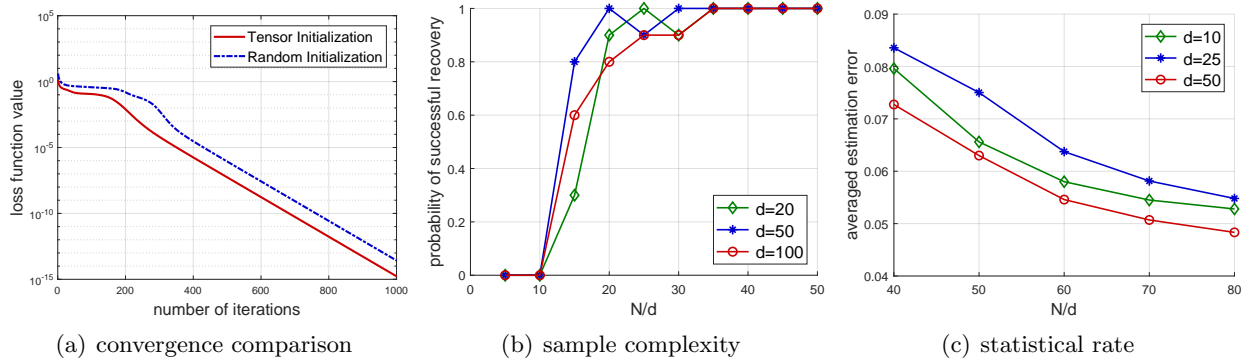

(a) convergence comparison     (b) sample complexity     (c) statistical rate

Figure 2: (a) Comparison of convergence rate for gradient descent based algorithm using different initialization procedures. Here, we set input dimension $d = 10$, sample size $N = 5000$ and number of neurons $K = 5$. (b) Plot of successful recovery probability versus the ratio between sample size and input dimension $N/d$, which illustrates that the sample complexity scales linearly with $d$. (c) Plot of averaged estimation error versus the rescaled sample size $N/d$ based on our method under different settings.

## 6 Experiments

In this section, simulation results on synthetic datasets are demonstrated to justify our theory. We sample the input data $\{\mathbf{x}_i\}_{i=1}^N$ from standard Gaussian distribution, and generate the output labels $\{y_i\}_{i=1}^N$ based on the teacher network (1.1). The number of neurons in the hidden layer is set as $K = 5$. We generate the underlying parameter matrix $\mathbf{W}^* \in \mathbb{R}^{d \times K}$ such that $\mathbf{W}^* = \mathbf{U}\mathbf{\Sigma}\mathbf{V}^\top$, where $\mathbf{U}, \mathbf{V}$ are the left and right singular matrices of a $d \times K$ standard Gaussian matrix, and $\mathbf{\Sigma}$ is a diagonal matrix. The smallest singular value of $\mathbf{W}^*$ is to be 1 and the largest one is set to be 2, and thus the condition number $\kappa = 2$.

To begin with, we study the convergence rate of our proposed Algorithm 1 in the noiseless case using different initialization procedures. In particular, we compare the tensor initialization algorithm proposed in Zhong et al. (2017) and random initialization procedure, where the initial estimator $\mathbf{W}^0$ is generated randomly from a standard Gaussian distribution. We choose the dimension $d$ as 10 and sample size $N$ as 5000. The step size $\eta$ is set to be 0.5. For each intermediate iterate $\mathbf{W}^t$ returned by Algorithm 1, we compute the empirical loss function value $\widehat{\mathcal{L}}_N(\mathbf{W}^t)$. The logarithm of the empirical loss function value is plotted in Figure 2(a) against the number of iterations. It can be seen that both initialization methods, followed by gradient descent, achieve linear rate of convergence after a certain number of iterations, but tensor initialization leads to faster convergence than random initialization at early stage.

Moreover, we investigate the sample complexity requirement of the gradient descent algorithm in the noiseless setting. In particular, we consider three cases: (i) $d = 20$; (ii) $d = 50$; (iii) $d = 100$. For each case, we vary the sample size $N$, and repeat Algorithm 1 for 10 trials. A trial is considered to be successful if there exists a permutation matrix[3] $\mathbf{M}_\pi \in \mathbb{R}^{K \times K}$ such that the returned estimator

---

[3] A permutation matrix is a square binary matrix that has exactly one entry of 1 in each row and each column and 0's elsewhere.

$\mathbf{W}^T$ satisfies

$$\|\mathbf{W}^T - \mathbf{W}^* \cdot \mathbf{M}_\pi\|_F / \|\mathbf{W}^*\|_F \leq 10^{-3}.$$

The results of successful recovery probability of $\mathbf{W}^*$ under different ratio $N/d$ are reported in Figure 2(b). It can be seen from the plot that the sample complexity required by tensor initialization followed by gradient descent for learning one-hidden-layer ReLU networks is linear in the dimension $d$, which is in agreement with our theory.

Finally, we study the statistical rate of our method in the noisy setting. In particular, we consider the following three cases: (i) $d = 10$, (ii) $d = 25$, (iii) $d = 50$. Each element of the noise vector $\boldsymbol{\epsilon} = [\epsilon_1, ..., \epsilon_N]^\top$ is generated independently from Gaussian distribution $\mathcal{N}(0, 0.1)$. We run Algorithm 1 with tensor initialization for each case over 10 trials, and report the averaged estimation error of the final output $\mathbf{W}^T$, i.e., $\|\mathbf{W}^T - \mathbf{W}^*\mathbf{M}_\pi\|_F$. Recall that $\mathbf{M}_\pi$ denotes the optimal permutation matrix with respect to $\mathbf{W}^T$ and $\mathbf{W}^*$. The results are displayed in Figure 2(c), which demonstrates that the averaged estimation error is well aligned with the rescaled sample size under different cases, which confirms that the statistical rate of the output of gradient descent for training one-hidden-layer ReLU networks is indeed in the order of $\widetilde{O}(d/N)$.

## 7 Conclusions and Future Work

In this paper, we studied the empirical risk minimization for training one-hidden-layer ReLU networks using gradient descent. We proved that gradient descent can converge to the ground-truth parameters at a linear rate up to some statistical error with sample complexity $\widetilde{O}(d)$. While the presented results are specific to shallow neural networks, we believe that they can shed light on understanding the learning of deep networks.

As for future work, one important but challenging direction is to study the global optimization landscape and learning guarantees for deeper neural networks with more than one hidden layers. Another future direction is to investigate whether the Gaussian input assumption can be relaxed to more general distribution assumption as done by Du et al. (2017a) for learning one convolutional filter. Last but not least, it would be interesting to study the theoretical guarantee for learning ReLU networks using random initialized gradient descent.

## A Proofs of Technical Lemmas in Section 5

### A.1 Proof of Lemma 5.1

To prove Lemma 5.1, we need to make use of the following auxiliary lemmas.

**Lemma A.1.** (Zhong et al., 2017) The Hessian of the population loss function $\mathcal{L}(\mathbf{W})$ at $\mathbf{W}^*$ satisfies
$$\nabla^2 \mathcal{L}(\mathbf{W}^*) \succeq 2\mu \mathbf{I},$$
where $\mu = c/(\lambda \kappa^2)$, and $c > 0$ is an absolute constant.

Lemma A.1 characterizes a lower bound of the Hessian $\nabla^2 \mathcal{L}(\mathbf{W}^*)$. Please refer to Lemma D.6 in Zhong et al. (2017) for the detailed proof.

**Lemma A.2.** (Safran and Shamir, 2017) The Hessian of the population loss function $\nabla^2 \mathcal{L}(\cdot)$ is $\rho$-Lipschitz within the region $\Omega = \{\mathbf{W} \in \mathbb{R}^{d \times K} \mid \|\mathbf{W} - \mathbf{W}^*\|_F \leq \sigma_K/2\}$, i.e., for any $\mathbf{W}_1, \mathbf{W}_2 \in \Omega$

$$\|\nabla^2 \mathcal{L}(\mathbf{W}_1) - \nabla^2 \mathcal{L}(\mathbf{W}_2)\|_2 \leq \rho \|\mathbf{W}_1 - \mathbf{W}_2\|_F,$$

where $\rho = c\kappa K^2/\sigma_K$, and $c > 0$ is a constant.

Lemma A.2 characterizes the Lipschitz property regarding the Hessian of the population loss function $\nabla^2 \mathcal{L}(\cdot)$ around $\mathbf{W}^*$, where the detailed proofs can be found in Lemma 6 in Safran and Shamir (2017). We provide the explicit calculation of the Hessian Lipschitz parameter $\rho$ in Appendix B.1.

Now we are ready to prove Lemma 5.1.

*Proof of Lemma 5.1.* According to Lemma A.2, we have

$$\|\nabla^2 \mathcal{L}(\mathbf{W}) - \nabla^2 \mathcal{L}(\mathbf{W}^*)\|_2 \leq c_1 \frac{\kappa K^2}{\sigma_K} \|\mathbf{W} - \mathbf{W}^*\|_F \leq c_2 \frac{1}{\lambda \kappa^2} \leq \mu,$$

where the first inequality is due to $\|\mathbf{W} - \mathbf{W}^*\|_F \leq c\sigma_K/(\lambda \kappa^3 K^2)$. Therefore, according to Lemma A.1 and Weyl's Lemma, we obtain

$$\sigma_{\min}\big(\nabla^2 \mathcal{L}(\mathbf{W})\big) \geq \sigma_{\min}\big(\nabla^2 \mathcal{L}(\mathbf{W}^*)\big) - \|\nabla^2 \mathcal{L}(\mathbf{W}) - \nabla^2 \mathcal{L}(\mathbf{W}^*)\|_2 \geq \mu,$$

where $\sigma_{\min}(\cdot)$ denotes the smallest eigenvalue. $\square$

### A.2 Proof of Lemma 5.2

*Proof of Lemma 5.2.* For any matrix $\mathbf{W} \in \mathbb{R}^{d \times K}$, we denote by $w_{\max} = \operatorname{argmax}_{i \in [K]} \|\mathbf{w}_i\|_2$ and $w_{\min}$ by $w_{\min} = \operatorname{argmin}_{i \in [K]} \|\mathbf{w}_i\|_2$, where $\mathbf{w}_i$ is the $i$-th column of $\mathbf{W}$. We also define $w_{\max}^* = \operatorname{argmax}_{i \in [K]} \|\mathbf{w}_i^*\|_2$, where $\mathbf{w}_i^*$ is the $i$-th column of $\mathbf{W}^*$. According to the definition of the spectral norm, we have

$$\sigma_K(\mathbf{W}) \leq \|\mathbf{W}\mathbf{e}_i\|_2 = \|\mathbf{w}_i\|_2 \leq \sigma_1(\mathbf{W}),$$

where $\mathbf{e}_i \in \mathbb{R}^k$ denotes the basis vector such that the $i$-th coordinate is 1 and others are 0. By assumption, we have $\|\mathbf{W} - \mathbf{W}^*\|_2 \leq \sigma_K/2$. Thus, by Weyl's Lemma, we further obtain $\sigma_1(\mathbf{W}) \leq 3\sigma_1/2$ and $\sigma_K(\mathbf{W}) \geq \sigma_K/2$. According to Lemma 7 in Safran and Shamir (2017), we have the following expression of the smoothness parameter $L$

$$L = c_1 K^2 \frac{\|w_{\max}\|_2}{\|w_{\min}\|_2} + c_2 K^2 \frac{\|w_{\max}^*\|_2}{\|w_{\min}\|_2},$$

where $c_1, c_2$ are absolute constants. This implies

$$L \leq c_3 \kappa K^2,$$

where $c_3$ is an absolute constant. $\square$

## A.3  Proof of Lemma 5.3

According to the expression of the gradient of the empirical loss function $\nabla \widehat{\mathcal{L}}_N(\mathbf{W})$ in (3.2), we have

$$\text{vec}(\nabla \widehat{\mathcal{L}}_N(\mathbf{W})) = \widehat{\boldsymbol{\Omega}}(\mathbf{W}, \mathbf{W})\text{vec}(\mathbf{W}) - \widehat{\boldsymbol{\Omega}}(\mathbf{W}^*, \mathbf{W})\text{vec}(\mathbf{W}^*) - \text{vec}(\mathbf{E}(\mathbf{W})), \qquad (A.1)$$

where $\widehat{\boldsymbol{\Omega}}(\mathbf{W}, \mathbf{W})$ and $\widehat{\boldsymbol{\Omega}}(\mathbf{W}^*, \mathbf{W})$ are defined as

$$\widehat{\boldsymbol{\Omega}}(\mathbf{W}, \mathbf{W}) = \begin{bmatrix} \widehat{\boldsymbol{\Sigma}}(\mathbf{w}_1, \mathbf{w}_1) & \dots & \widehat{\boldsymbol{\Sigma}}(\mathbf{w}_K, \mathbf{w}_1) \\ \vdots & \ddots & \vdots \\ \widehat{\boldsymbol{\Sigma}}(\mathbf{w}_1, \mathbf{w}_K) & \dots & \widehat{\boldsymbol{\Sigma}}(\mathbf{w}_K, \mathbf{w}_K) \end{bmatrix} \quad \text{and} \quad \widehat{\boldsymbol{\Omega}}(\mathbf{W}^*, \mathbf{W}) = \begin{bmatrix} \widehat{\boldsymbol{\Sigma}}(\mathbf{w}_1^*, \mathbf{w}_1) & \dots & \widehat{\boldsymbol{\Sigma}}(\mathbf{w}_K^*, \mathbf{w}_1) \\ \vdots & \ddots & \vdots \\ \widehat{\boldsymbol{\Sigma}}(\mathbf{w}_1^*, \mathbf{w}_K) & \dots & \widehat{\boldsymbol{\Sigma}}(\mathbf{w}_K^*, \mathbf{w}_K) \end{bmatrix},$$

and the error matrix $\mathbf{E}(\mathbf{W})$ is defined as

$$\mathbf{E}(\mathbf{W}) = \frac{1}{N} \sum_{i \in N} \epsilon_i \mathbf{x}_i \cdot \begin{bmatrix} \mathbb{1}\{\mathbf{w}_1^\top \mathbf{x}_i \geq 0\} & \cdots & \mathbb{1}\{\mathbf{w}_K^\top \mathbf{x}_i \geq 0\} \end{bmatrix}.$$

Here, $\widehat{\boldsymbol{\Sigma}}(\mathbf{w}_j, \mathbf{w}_k)$ and $\widehat{\boldsymbol{\Sigma}}(\mathbf{w}_j^*, \mathbf{w}_k)$ are defined in (3.3). Similarly, we obtain that the gradient of the population loss function $\mathcal{L}(\mathbf{W})$ as in (3.1) satisfies

$$\text{vec}(\nabla \mathcal{L}(\mathbf{W})) = \boldsymbol{\Omega}(\mathbf{W}, \mathbf{W})\text{vec}(\mathbf{W}) - \boldsymbol{\Omega}(\mathbf{W}^*, \mathbf{W})\text{vec}(\mathbf{W}^*), \qquad (A.2)$$

where $\boldsymbol{\Omega}(\mathbf{W}, \mathbf{W})$ and $\boldsymbol{\Omega}(\mathbf{W}^*, \mathbf{W})$ are defined as

$$\boldsymbol{\Omega}(\mathbf{W}, \mathbf{W}) = \begin{bmatrix} \boldsymbol{\Sigma}(\mathbf{w}_1, \mathbf{w}_1) & \dots & \boldsymbol{\Sigma}(\mathbf{w}_K, \mathbf{w}_1) \\ \vdots & \ddots & \vdots \\ \boldsymbol{\Sigma}(\mathbf{w}_1, \mathbf{w}_K) & \dots & \boldsymbol{\Sigma}(\mathbf{w}_K, \mathbf{w}_K) \end{bmatrix} \quad \text{and} \quad \boldsymbol{\Omega}(\mathbf{W}^*, \mathbf{W}) = \begin{bmatrix} \boldsymbol{\Sigma}(\mathbf{w}_1^*, \mathbf{w}_1) & \dots & \boldsymbol{\Sigma}(\mathbf{w}_K^*, \mathbf{w}_1) \\ \vdots & \ddots & \vdots \\ \boldsymbol{\Sigma}(\mathbf{w}_1^*, \mathbf{w}_K) & \dots & \boldsymbol{\Sigma}(\mathbf{w}_K^*, \mathbf{w}_K) \end{bmatrix}.$$

Here, we define $\boldsymbol{\Sigma}(\mathbf{w}_j, \mathbf{w}_k)$ and $\boldsymbol{\Sigma}(\mathbf{w}_j^*, \mathbf{w}_k)$ as

$$\boldsymbol{\Sigma}(\mathbf{w}_j, \mathbf{w}_k) = \mathbb{E}_{\boldsymbol{X} \sim \mathcal{D}_X}\left[ \boldsymbol{X}\boldsymbol{X}^\top \cdot \mathbb{1}\{\mathbf{w}_j^\top \boldsymbol{X} \geq 0\} \cdot \mathbb{1}\{\mathbf{w}_k^\top \boldsymbol{X} \geq 0\} \right],$$

$$\boldsymbol{\Sigma}(\mathbf{w}_j^*, \mathbf{w}_k) = \mathbb{E}_{\boldsymbol{X} \sim \mathcal{D}_X}\left[ \boldsymbol{X}\boldsymbol{X}^\top \cdot \mathbb{1}\{\mathbf{w}_j^{*\top} \boldsymbol{X} \geq 0\} \cdot \mathbb{1}\{\mathbf{w}_k^\top \boldsymbol{X} \geq 0\} \right].$$

In order to prove Lemma 5.3, we need the following two results.

**Lemma A.3.** For all $\mathbf{W} \in \mathbb{R}^{d \times K}$ such that $\|\mathbf{W} - \mathbf{W}^*\|_F \leq c\sigma_K/(\lambda \kappa^3 K^2)$, where $c$ is an absolute constant, there exist absolute constants $c_1$ such that

$$\|\widehat{\boldsymbol{\Omega}}(\mathbf{W}^*, \mathbf{W}) - \boldsymbol{\Omega}(\mathbf{W}^*, \mathbf{W})\|_2 \leq c_1 K^{5/2} \sqrt{\frac{d \log N}{N}}$$

holds with probability at least $1 - d^{-10}$.

**Lemma A.4.** For all $\mathbf{W} \in \mathbb{R}^{d \times K}$ such that $\|\mathbf{W} - \mathbf{W}^*\|_F \leq c\sigma_K/(\lambda \kappa^3 K^2)$, where $c$ is an absolute constant, there exist absolute constants $c_1$ such that

$$\|\mathbf{E}(\mathbf{W})\|_F \leq c_1 \nu K^{3/2} \sqrt{\frac{d \log N}{N}}$$

holds with probability at least $1 - d^{-10}$.

*Proof of Lemma 5.3.* According to (A.1) and (A.2), we have

$$\|\nabla\widehat{\mathcal{L}}_N(\mathbf{W}) - \nabla\mathcal{L}(\mathbf{W})\|_F \leq \underbrace{\|\widehat{\mathbf{\Omega}}(\mathbf{W},\mathbf{W}) - \mathbf{\Omega}(\mathbf{W},\mathbf{W})\|_2}_{I_1} \cdot \|\mathbf{W}\|_F + \underbrace{\|\widehat{\mathbf{\Omega}}(\mathbf{W}^*,\mathbf{W}) - \mathbf{\Omega}(\mathbf{W}^*,\mathbf{W})\|_2}_{I_2} \cdot \|\mathbf{W}^*\|_F$$
$$+ \underbrace{\|\mathbf{E}(\mathbf{W})\|_F}_{I_3}. \quad\quad\quad (A.3)$$

**Bounding $I_1$ and $I_2$:** According to Lemma A.3, we can obtain

$$\|\widehat{\mathbf{\Omega}}(\mathbf{W}^*,\mathbf{W}) - \mathbf{\Omega}(\mathbf{W}^*,\mathbf{W})\|_2 \leq c_1 K^{5/2}\sqrt{\frac{d\log N}{N}} \quad\quad (A.4)$$

holds with probability at least $1 - 1/d^{10}$. Similarly, we can obtain the same high probability bound for $\|\widehat{\mathbf{\Omega}}(\mathbf{W},\mathbf{W}) - \mathbf{\Omega}(\mathbf{W},\mathbf{W})\|_2$.

**Bounding $I_3$:** Now we turn to bound the Frobenius norm of the error matrix $\mathbf{E}(\mathbf{W})$. According to Lemma A.4, we have

$$\|\mathbf{E}(\mathbf{W})\|_F \leq c_2 \nu K^{3/2}\sqrt{\frac{d\log N}{N}} \quad\quad (A.5)$$

holds with probability at least $1 - 1/d^{10}$.

Finally, plugging (A.4) and (A.5) into (A.3), we have

$$\|\nabla\widehat{\mathcal{L}}_N(\mathbf{W}) - \nabla\mathcal{L}(\mathbf{W})\|_F \leq c_3 K^{5/2}\sqrt{\frac{d\log N}{N}}(\|\mathbf{W}\|_F + \|\mathbf{W}^*\|_F + \nu)$$
$$\leq 3c_3 K^{5/2}\sqrt{\frac{d\log N}{N}}(\|\mathbf{W}^*\|_F + \nu)$$

holds with high probability, where the last inequality is due to $\|\mathbf{W} - \mathbf{W}^*\|_F \leq \sigma_K/2 \leq \|\mathbf{W}^*\|_F/2$. $\square$

# B  Proof of Auxiliary Lemma in Appendix A

## B.1  Proof of Lemma A.2

*Proof.* Similar to previous analysis in Lemma 5.2, we can derive that

$$\sigma_K(\mathbf{W}) \leq \|\mathbf{W}\mathbf{e}_i\|_2 = \|\mathbf{w}_i\|_2 \leq \sigma_1(\mathbf{W}),$$

where $\mathbf{e}_i \in \mathbb{R}^k$ denotes the basis vector such that the $i$-th coordinate is one and others are zero. Since $\|\mathbf{W} - \mathbf{W}^*\|_2 \leq \sigma_K/2$, by Weyl's Lemma, we have $\sigma_1(\mathbf{W}) \leq 3\sigma_1/2$ and $\sigma_K(\mathbf{W}) \geq \sigma_K/2$. According to Lemma 6 in Safran and Shamir (2017), we have the following expression of the Hessian Lipschitz parameter $\rho$

$$\rho = c_1 K^2 \frac{\|w_{\max}\|_2 + \|w_{\min}\|_2 + \|w^*_{\max}\|_2}{\|w_{\min}\|_2^2} \leq c_2 \frac{\kappa K^2}{\sigma_K},$$

where $c_1, c_2$ are absolute constants. $\square$

## B.2 Proof of Lemma A.3

*Proof.* The proof of this lemma is inspired by the proof of Theorem 1 in Mei et al. (2016) and Lemma 3 in Fu et al. (2018). Let $\mathcal{N}_\zeta$ be the $\zeta$-covering of the ball $\mathcal{B}_r(\mathbf{W}^*)$. Let $\widetilde{\mathbf{W}} = \operatorname{argmin}_{\mathbf{W}' \in \mathcal{N}_\zeta} \|\mathbf{W}' - \mathbf{W}\|_F \leq \zeta$ for all $\mathbf{W} \in \mathcal{B}_r(\mathbf{W}^*)$. Since we have

$$\widehat{\mathbf{\Omega}}(\mathbf{W}^*, \mathbf{W}) - \mathbf{\Omega}(\mathbf{W}^*, \mathbf{W}) = \widehat{\mathbf{\Omega}}(\mathbf{W}^*, \mathbf{W}) - \widehat{\mathbf{\Omega}}(\mathbf{W}^*, \widetilde{\mathbf{W}}) + \widehat{\mathbf{\Omega}}(\mathbf{W}^*, \widetilde{\mathbf{W}}) - \mathbf{\Omega}(\mathbf{W}^*, \widetilde{\mathbf{W}}) \\ + \mathbf{\Omega}(\mathbf{W}^*, \widetilde{\mathbf{W}}) - \mathbf{\Omega}(\mathbf{W}^*, \mathbf{W}),$$

which implies

$$\|\widehat{\mathbf{\Omega}}(\mathbf{W}^*, \mathbf{W}) - \mathbf{\Omega}(\mathbf{W}^*, \mathbf{W})\|_2 \leq \|\widehat{\mathbf{\Omega}}(\mathbf{W}^*, \mathbf{W}) - \widehat{\mathbf{\Omega}}(\mathbf{W}^*, \widetilde{\mathbf{W}})\|_2 + \|\widehat{\mathbf{\Omega}}(\mathbf{W}^*, \widetilde{\mathbf{W}}) - \mathbf{\Omega}(\mathbf{W}^*, \widetilde{\mathbf{W}})\|_2 \\ + \|\mathbf{\Omega}(\mathbf{W}^*, \widetilde{\mathbf{W}}) - \mathbf{\Omega}(\mathbf{W}^*, \mathbf{W})\|_2.$$

Therefore, we can obtain

$$\mathbb{P}\left(\sup_{\mathbf{W} \in \mathcal{B}_r(\mathbf{W}^*)} \|\widehat{\mathbf{\Omega}}(\mathbf{W}^*, \mathbf{W}) - \mathbf{\Omega}(\mathbf{W}^*, \mathbf{W})\|_2 \geq t\right) \leq \mathbb{P}\left(\sup_{\mathbf{W} \in \mathcal{B}_r(\mathbf{W}^*)} \|\widehat{\mathbf{\Omega}}(\mathbf{W}^*, \mathbf{W}) - \widehat{\mathbf{\Omega}}(\mathbf{W}^*, \widetilde{\mathbf{W}})\|_2 \geq \frac{t}{3}\right) \\ + \mathbb{P}\left(\sup_{\widetilde{\mathbf{W}} \in \mathcal{N}_\zeta} \|\widehat{\mathbf{\Omega}}(\mathbf{W}^*, \widetilde{\mathbf{W}}) - \mathbf{\Omega}(\mathbf{W}^*, \widetilde{\mathbf{W}})\|_2 \geq \frac{t}{3}\right) \\ + \mathbb{P}\left(\sup_{\mathbf{W} \in \mathcal{B}_r(\mathbf{W}^*)} \|\mathbf{\Omega}(\mathbf{W}^*, \widetilde{\mathbf{W}}) - \mathbf{\Omega}(\mathbf{W}^*, \mathbf{W})\|_2 \geq \frac{t}{3}\right).$$

**Bounding** $\sup_{\mathbf{W} \in \mathcal{B}_r(\mathbf{W}^*)} \|\mathbf{\Omega}(\mathbf{W}^*, \widetilde{\mathbf{W}}) - \mathbf{\Omega}(\mathbf{W}^*, \mathbf{W})\|_2$: Based on the definition of spectral norm, we have

$$\|\mathbf{\Omega}(\mathbf{W}^*, \widetilde{\mathbf{W}}) - \mathbf{\Omega}(\mathbf{W}^*, \mathbf{W})\|_2 = \sup_{\|\mathbf{U}\|_F=1, \|\widetilde{\mathbf{U}}\|_F=1} \operatorname{vec}(\widetilde{\mathbf{U}})^\top (\mathbf{\Omega}(\mathbf{W}^*, \widetilde{\mathbf{W}}) - \mathbf{\Omega}(\mathbf{W}^*, \mathbf{W})) \operatorname{vec}(\mathbf{U})$$

$$= \sup_{\|\mathbf{U}\|_F=1, \|\widetilde{\mathbf{U}}\|_F=1} \sum_{(j,k) \in [K] \times [K]} \widetilde{\mathbf{u}}_j^\top \left[\mathbf{\Sigma}(\mathbf{w}_j^*, \widetilde{\mathbf{w}}_k) - \mathbf{\Sigma}(\mathbf{w}_j^*, \mathbf{w}_k)\right] \mathbf{u}_k$$

$$\leq \sum_{(j,k) \in [K] \times [K]} \sup_{\|\widetilde{\mathbf{u}}_j\|_2=1, \|\mathbf{u}_k\|_2=1} \widetilde{\mathbf{u}}_j^\top \left[\mathbf{\Sigma}(\mathbf{w}_j^*, \widetilde{\mathbf{w}}_k) - \mathbf{\Sigma}(\mathbf{w}_j^*, \mathbf{w}_k)\right] \mathbf{u}_k$$

$$= \sum_{(j,k) \in [K] \times [K]} \|\mathbf{\Sigma}(\mathbf{w}_j^*, \widetilde{\mathbf{w}}_k) - \mathbf{\Sigma}(\mathbf{w}_j^*, \mathbf{w}_k)\|_2. \quad \text{(B.1)}$$

In addition, we have

$$\mathbf{\Sigma}(\mathbf{w}_j^*, \widetilde{\mathbf{w}}_k) - \mathbf{\Sigma}(\mathbf{w}_j^*, \mathbf{w}_k) = \mathbb{E}_{\mathbf{X} \sim \mathcal{D}_X}\left[\mathbf{X}\mathbf{X}^\top \cdot \left(\mathbb{1}\{\widetilde{\mathbf{w}}_k^\top \mathbf{X} \geq 0\} - \mathbb{1}\{\mathbf{w}_k^\top \mathbf{X} \geq 0\}\right) \cdot \mathbb{1}\{\mathbf{w}_j^{*\top} \mathbf{X} \geq 0\}\right].$$

Thus we have

$$\|\mathbf{\Sigma}(\mathbf{w}_j^*, \widetilde{\mathbf{w}}_k) - \mathbf{\Sigma}(\mathbf{w}_j^*, \mathbf{w}_k)\|_2 \leq d \cdot \angle \widetilde{\mathbf{w}}_k, \mathbf{w}_k$$

$$\leq d \cdot \arcsin\left(\frac{\|\widetilde{\mathbf{w}}_k - \mathbf{w}_k\|_2}{\|\mathbf{w}_k\|_2}\right)$$

$$\leq d \cdot \arcsin\left(\frac{\zeta}{\|\mathbf{w}_k^*\|_2 - r}\right)$$

$$\leq 2d\tau\zeta,$$

where $\tau = \max_l \left(1/(\|\mathbf{w}_l^*\|_2 - r)\right)$, the first inequality is due to Lemma C.7 with $\angle \widetilde{\mathbf{w}}_k, \mathbf{w}_k$ representing the angle between $\widetilde{\mathbf{w}}_k$ and $\mathbf{w}_k$, the third inequality is due to the definition of $\widetilde{\mathbf{W}}$, and the last inequality holds due to the fact that $\arcsin x \leq 2x$ when $x \in [0, \pi/4]$. According to (B.1), we have

$$\sup_{\mathbf{W} \in \mathcal{B}_r(\mathbf{W}^*)} \|\mathbf{\Omega}(\mathbf{W}^*, \widetilde{\mathbf{W}}) - \mathbf{\Omega}(\mathbf{W}^*, \mathbf{W})\|_2 \leq c_1 dK^2 \zeta. \tag{B.2}$$

Therefore, as long as $t > 3c_1 dK^2 \zeta$, we have

$$\mathbb{P}\left(\sup_{\mathbf{W} \in \mathcal{B}_r(\mathbf{W}^*)} \|\mathbf{\Omega}(\mathbf{W}^*, \widetilde{\mathbf{W}}) - \mathbf{\Omega}(\mathbf{W}^*, \mathbf{W})\|_2 \geq \frac{t}{3}\right) = 0. \tag{B.3}$$

**Bounding** $\sup_{\widetilde{\mathbf{W}} \in \mathcal{N}_\zeta} \|\widehat{\mathbf{\Omega}}(\mathbf{W}^*, \widetilde{\mathbf{W}}) - \mathbf{\Omega}(\mathbf{W}^*, \widetilde{\mathbf{W}})\|_2$: Similar to (B.1), we have

$$\|\widehat{\mathbf{\Omega}}(\mathbf{W}^*, \widetilde{\mathbf{W}}) - \mathbf{\Omega}(\mathbf{W}^*, \widetilde{\mathbf{W}})\|_2 \leq \sum_{(j,k) \in [K] \times [K]} \|\widehat{\mathbf{\Sigma}}(\mathbf{w}_j^*, \widetilde{\mathbf{w}}_k) - \mathbf{\Sigma}(\mathbf{w}_j^*, \widetilde{\mathbf{w}}_k)\|_2.$$

In addition, for any $(j, k) \in [K] \times [K]$, we have

$$\|\widehat{\mathbf{\Sigma}}_2(\mathbf{w}_j^*, \widetilde{\mathbf{w}}_k) - \mathbf{\Sigma}_2(\mathbf{w}_j^*, \widetilde{\mathbf{w}}_k)\|_2$$
$$= \left\| \frac{1}{N} \sum_{i=1}^N \mathbf{x}_i \mathbf{x}_i^\top \cdot \mathbb{1}\{\mathbf{w}_j^{*\top} \mathbf{x}_i \geq 0, \widetilde{\mathbf{w}}_k^\top \mathbf{x}_i \geq 0\} - \mathbb{E}_{\mathbf{X} \sim \mathcal{D}_X}\left[\mathbf{X}\mathbf{X}^\top \cdot \mathbb{1}\{\mathbf{w}_j^{*\top} \mathbf{X} \geq 0, \widetilde{\mathbf{w}}_k^\top \mathbf{X} \geq 0\}\right] \right\|_2$$
$$= \left\| \frac{1}{N} \sum_{i=1}^N \widetilde{\mathbf{x}}_i \widetilde{\mathbf{x}}_i^\top - \mathbb{E}_{\mathbf{X} \sim \mathcal{D}_X}\left[\widetilde{\mathbf{X}} \widetilde{\mathbf{X}}^\top\right] \right\|_2,$$

where we denote $\widetilde{\mathbf{x}}_i = \mathbf{x}_i \cdot \mathbb{1}\{\mathbf{w}_j^{*\top} \mathbf{x}_i \geq 0, \widetilde{\mathbf{w}}_k^\top \mathbf{x}_i \geq 0\}$, $\widetilde{\mathbf{X}} = \mathbf{X} \cdot \mathbb{1}\{\mathbf{w}_j^{*\top} \mathbf{X} \geq 0, \widetilde{\mathbf{w}}_k^\top \mathbf{X} \geq 0\}$.

We claim that for any fixed $\mathbf{w}_j^*, \widetilde{\mathbf{w}}_k \in \mathbb{R}^d$, $\widetilde{\mathbf{X}} \in \mathbb{R}^d$ follows sub-Gaussian distribution with sub-Gaussian norm $C$, where $C$ is an absolute constant. To prove this claim, we make use of the definition of sub-Gaussian random vector (See Appendix C for the formal definition of sub-Gaussian random vector). Note that $\mathbf{X} \in \mathbb{R}^d$ follows standard Gaussian distribution, thus for all $\boldsymbol{\alpha} \in S^{d-1}$ and $p \geq 1$, we have

$$\left(\mathbb{E}_{\mathbf{X} \sim \mathcal{D}_X} |\boldsymbol{\alpha}^\top \widetilde{\mathbf{X}}|^p\right)^{1/p} = \left(\int_{\mathbf{w}_j^\top \mathbf{Z} \geq 0, \mathbf{w}_k^\top \mathbf{Z} \geq 0} |\boldsymbol{\alpha}^\top \mathbf{Z}|^p f_{\mathbf{X}}(\mathbf{Z}) \, d\mathbf{Z}\right)^{1/p}$$
$$\leq \left(\mathbb{E}_{\mathbf{X} \sim \mathcal{D}_X} |\boldsymbol{\alpha}^\top \mathbf{X}|^p\right)^{1/p} \leq \|\boldsymbol{\alpha}^\top \mathbf{X}\|_{\psi_2} \cdot \sqrt{p} \leq C\sqrt{p},$$

where $f_{\mathbf{X}}(\cdot)$ denotes the probability density function of $\mathbf{X}$. By definition, we complete the proof of the claim. We further obtain that $\widetilde{\mathbf{x}}_i$'s are sampled from i.i.d. sub-Gaussian distribution with sub-Gaussian norm $C$. By the union bound, we have

$$\mathbb{P}\left(\sup_{\widetilde{\mathbf{W}} \in \mathcal{N}_\zeta} \|\widehat{\mathbf{\Omega}}(\mathbf{W}^*, \widetilde{\mathbf{W}}) - \mathbf{\Omega}(\mathbf{W}^*, \widetilde{\mathbf{W}})\|_2 \geq \frac{t}{3}\right) \leq |\mathcal{N}_\zeta| \mathbb{P}\left(\|\widehat{\mathbf{\Omega}}(\mathbf{W}^*, \widetilde{\mathbf{W}}) - \mathbf{\Omega}(\mathbf{W}^*, \widetilde{\mathbf{W}})\|_2 \geq \frac{t}{3}\right)$$
$$= |\mathcal{N}_\zeta| \mathbb{P}\left(\sum_{(j,k) \in [K] \times [K]} \|\widehat{\mathbf{\Sigma}}(\mathbf{w}_j^*, \widetilde{\mathbf{w}}_k) - \mathbf{\Sigma}(\mathbf{w}_j^*, \widetilde{\mathbf{w}}_k)\|_2 \geq \frac{t}{3}\right)$$
$$\leq |\mathcal{N}_\zeta| K^2 \mathbb{P}\left(\|\widehat{\mathbf{\Sigma}}(\mathbf{w}_j^*, \widetilde{\mathbf{w}}_k) - \mathbf{\Sigma}(\mathbf{w}_j^*, \widetilde{\mathbf{w}}_k)\|_2 \geq \frac{t}{3K^2}\right).$$

In addition, according to Lemma C.3, we can obtain that

$$\mathbb{P}\bigg(\|\widehat{\boldsymbol{\Sigma}}(\mathbf{w}_j^*, \widetilde{\mathbf{w}}_k) - \boldsymbol{\Sigma}(\mathbf{w}_j^*, \widetilde{\mathbf{w}}_k)\|_2 \geq \frac{t}{3K^2}\bigg) \leq 2\exp(-c_2 N t^2/K^4),$$

which implies that

$$\mathbb{P}\bigg(\sup_{\widetilde{\mathbf{W}} \in \mathcal{N}_\zeta} \|\widehat{\boldsymbol{\Omega}}(\mathbf{W}^*, \widetilde{\mathbf{W}}) - \boldsymbol{\Omega}(\mathbf{W}^*, \widetilde{\mathbf{W}})\|_2 \geq \frac{t}{3}\bigg) \leq 2K^2 \exp(dK \log(3r/\zeta) - c_2 N t^2/K^4),$$

where the inequality comes from Lemma C.5 such that $\log |\mathcal{N}_\zeta| \leq dK \log(3r/\zeta)$. Hence if we choose $t \geq c_3 \sqrt{(dK^5 \log(3r/\zeta) + K^4 \log(4K^2/\delta))/N}$, we have

$$\mathbb{P}\bigg(\sup_{\widetilde{\mathbf{W}} \in \mathcal{N}_\zeta} \|\widehat{\boldsymbol{\Omega}}(\mathbf{W}^*, \widetilde{\mathbf{W}}) - \boldsymbol{\Omega}(\mathbf{W}^*, \widetilde{\mathbf{W}})\|_2 \geq \frac{t}{3}\bigg) \leq \frac{\delta}{2}. \tag{B.4}$$

**Bounding $\sup_{\mathbf{W} \in \mathcal{B}_r(\mathbf{W}^*)} \|\widehat{\boldsymbol{\Omega}}(\mathbf{W}^*, \mathbf{W}) - \widehat{\boldsymbol{\Omega}}(\mathbf{W}^*, \widetilde{\mathbf{W}})\|_2$:** Similar to (B.1), we have

$$\|\widehat{\boldsymbol{\Omega}}(\mathbf{W}^*, \mathbf{W}) - \widehat{\boldsymbol{\Omega}}(\mathbf{W}^*, \widetilde{\mathbf{W}})\|_2 \leq \sum_{(j,k) \in [K] \times [K]} \|\widehat{\boldsymbol{\Sigma}}(\mathbf{w}_j^*, \mathbf{w}_k) - \widehat{\boldsymbol{\Sigma}}(\mathbf{w}_j^*, \widetilde{\mathbf{w}}_k)\|_2,$$

which implies

$$\sup_{\mathbf{W} \in \mathcal{B}_r(\mathbf{W}^*)} \|\widehat{\boldsymbol{\Omega}}(\mathbf{W}^*, \mathbf{W}) - \widehat{\boldsymbol{\Omega}}(\mathbf{W}^*, \widetilde{\mathbf{W}})\|_2 \leq \sum_{(j,k) \in [K] \times [K]} \sup_{\mathbf{W} \in \mathcal{B}_r(\mathbf{W}^*)} \|\widehat{\boldsymbol{\Sigma}}(\mathbf{w}_j^*, \mathbf{w}_k) - \widehat{\boldsymbol{\Sigma}}(\mathbf{w}_j^*, \widetilde{\mathbf{w}}_k)\|_2.$$

Since we have

$$\widehat{\boldsymbol{\Sigma}}(\mathbf{w}_j^*, \mathbf{w}_k) - \widehat{\boldsymbol{\Sigma}}(\mathbf{w}_j^*, \widetilde{\mathbf{w}}_k) = \frac{1}{N} \sum_{i=1}^N \mathbf{x}_i \mathbf{x}_i^\top \cdot \big(\mathbb{1}\{\mathbf{w}_k^\top \mathbf{x}_i \geq 0\} - \mathbb{1}\{\widetilde{\mathbf{w}}_k^\top \mathbf{x}_i \geq 0\}\big) \cdot \mathbb{1}\{\mathbf{w}_j^{*\top} \mathbf{x}_i \geq 0\},$$

we can obtain

$$\sup_{\mathbf{W} \in \mathcal{B}_r(\mathbf{W}^*)} \|\widehat{\boldsymbol{\Sigma}}(\mathbf{w}_j^*, \mathbf{w}_k) - \widehat{\boldsymbol{\Sigma}}(\mathbf{w}_j^*, \widetilde{\mathbf{w}}_k)\|_2 \leq \frac{1}{N} \sup_{\mathbf{W} \in \mathcal{B}_r(\mathbf{W}^*)} \bigg\|\sum_{i=1}^N \mathbf{x}_i \mathbf{x}_i^\top \cdot \mathbb{1}\{\mathbf{w}_k^\top \mathbf{x}_i \geq 0\} \mathbb{1}\{\widetilde{\mathbf{w}}_k^\top \mathbf{x}_i \leq 0\}\bigg\|_2$$

$$+ \frac{1}{N} \sup_{\mathbf{W} \in \mathcal{B}_r(\mathbf{W}^*)} \bigg\|\sum_{i=1}^N \mathbf{x}_i \mathbf{x}_i^\top \cdot \mathbb{1}\{\mathbf{w}_k^\top \mathbf{x}_i \leq 0\} \mathbb{1}\{\widetilde{\mathbf{w}}_k^\top \mathbf{x}_i \geq 0\}\bigg\|_2,$$

which implies

$$\sup_{\mathbf{W} \in \mathcal{B}_r(\mathbf{W}^*)} \|\widehat{\boldsymbol{\Sigma}}(\mathbf{w}_j^*, \mathbf{w}_k) - \widehat{\boldsymbol{\Sigma}}(\mathbf{w}_j^*, \widetilde{\mathbf{w}}_k)\|_2 \leq \max_i \|\mathbf{x}_i \mathbf{x}_i^\top\|_2 \cdot \sup_{\mathbf{W} \in \mathcal{B}_r(\mathbf{W}^*)} \mathbb{1}\{\mathbf{w}_k^\top \mathbf{x}_i \geq 0\} \mathbb{1}\{\widetilde{\mathbf{w}}_k^\top \mathbf{x}_i \leq 0\}$$

$$+ \max_i \|\mathbf{x}_i \mathbf{x}_i^\top\|_2 \cdot \sup_{\mathbf{W} \in \mathcal{B}_r(\mathbf{W}^*)} \mathbb{1}\{\mathbf{w}_k^\top \mathbf{x}_i \leq 0\} \mathbb{1}\{\widetilde{\mathbf{w}}_k^\top \mathbf{x}_i \geq 0\}.$$

In addition, by Markov inequality, we have

$$\mathbb{P}\bigg(\sup_{\mathbf{W} \in \mathcal{B}_r(\mathbf{W}^*)} \mathbb{1}\{\mathbf{w}_k^\top \mathbf{x}_i \geq 0\} \mathbb{1}\{\widetilde{\mathbf{w}}_k^\top \mathbf{x}_i \leq 0\} \geq t_1\bigg) \leq \frac{1}{t_1} \mathbb{E}_{\mathbf{X} \sim \mathcal{D}_X}\bigg[\sup_{\mathbf{W} \in \mathcal{B}_r(\mathbf{W}^*)} \mathbb{1}\{\mathbf{w}_k^\top \mathbf{x}_i \geq 0\} \mathbb{1}\{\widetilde{\mathbf{w}}_k^\top \mathbf{x}_i \leq 0\}\bigg].$$

According to the definition of $\widetilde{\mathbf{W}}$, we have

$$\mathbb{E}_{\boldsymbol{X} \sim \mathcal{D}_X} \left[ \sup_{\mathbf{W} \in \mathcal{B}_r(\mathbf{W}^*)} \mathbb{1}\{\mathbf{w}_k^\top \mathbf{x}_i \geq 0\} \mathbb{1}\{\widetilde{\mathbf{w}}_k^\top \mathbf{x}_i \leq 0\} \right] \leq \frac{1}{2\pi} \arcsin \frac{\zeta}{\|\mathbf{w}_k^*\|_2 - r} \leq \frac{\zeta \tau}{\pi},$$

where $\tau = \max_l \left(1/(\|\mathbf{w}_l^*\|_2 - r)\right)$, the last inequality is due to the fact that $\arcsin x \leq 2x$ if $x \in [0, \pi/4]$. Thus we have

$$\mathbb{P}\left( \sup_{\mathbf{W} \in \mathcal{B}_r(\mathbf{W}^*)} \mathbb{1}\{\mathbf{w}_k^\top \mathbf{x}_i \geq 0\} \mathbb{1}\{\widetilde{\mathbf{w}}_k^\top \mathbf{x}_i \leq 0\} \geq t_1 \right) \leq c_4 \frac{\zeta}{t_1}. \tag{B.5}$$

In addition, we have

$$\mathbb{P}\left( \max_i \|\mathbf{x}_i \mathbf{x}_i^\top\|_2 \geq t_2 \right) \leq 2N \exp(-c_5 t_2^2).$$

Thus we can obtain

$$\mathbb{P}\left( \max_i \|\mathbf{x}_i \mathbf{x}_i^\top\|_2 \cdot \sup_{\mathbf{W} \in \mathcal{B}_r(\mathbf{W}^*)} \mathbb{1}\{\mathbf{w}_k^\top \mathbf{x}_i \geq 0\} \mathbb{1}\{\widetilde{\mathbf{w}}_k^\top \mathbf{x}_i \leq 0\} \geq t_1 t_2 \right) \leq 2N \exp(-c_5 t_2^2) + c_4 \frac{\zeta}{t_1}.$$

Therefore, we can obtain

$$\mathbb{P}\left( \sup_{\mathbf{W} \in \mathcal{B}_r(\mathbf{W}^*)} \|\widehat{\boldsymbol{\Sigma}}(\mathbf{w}_j^*, \mathbf{w}_k) - \widehat{\boldsymbol{\Sigma}}(\mathbf{w}_j^*, \widetilde{\mathbf{w}}_k)\|_2 \geq 2t_1 t_2 \right) \leq 4N \exp(-c_5 t_2^2) + 2c_4 \frac{\zeta}{t_1}.$$

As a result, we have

$$\mathbb{P}\left( \sup_{\mathbf{W} \in \mathcal{B}_r(\mathbf{W}^*)} \|\widehat{\boldsymbol{\Omega}}(\mathbf{W}^*, \mathbf{W}) - \widehat{\boldsymbol{\Omega}}(\mathbf{W}^*, \widetilde{\mathbf{W}})\|_2 \geq 2K^2 t_1 t_2 \right) \leq 4NK^2 \exp(-c_5 t_2^2) + 2c_4 \frac{\zeta K^2}{t_1}.$$

Therefore, let $t_1 \geq c_6 \sqrt{\log(NK^2/\delta)}$, $t_2 \geq c_7 \zeta K^2/\delta$, and replace $2K^2 t_1 t_2$ with $t/3$, we can obtain that as long as $t \geq c_8 K^4 \zeta \sqrt{\log(NK^2/\delta)}/\delta$, the following holds

$$\mathbb{P}\left( \sup_{\mathbf{W} \in \mathcal{B}_r(\mathbf{W}^*)} \|\widehat{\boldsymbol{\Omega}}(\mathbf{W}^*, \mathbf{W}) - \widehat{\boldsymbol{\Omega}}(\mathbf{W}^*, \widetilde{\mathbf{W}})\|_2 \geq \frac{t}{3} \right) \leq \frac{\delta}{2}. \tag{B.6}$$

Therefore, according to (B.3), (B.4), and (B.6), if we choose $\delta = d^{-10}$, $\zeta = c_9 r \delta/N$, we have $t \geq c_{10} K^{5/2} \sqrt{d \log N/N}$, and we can obtain that for all $\mathbf{W} \in \mathcal{B}_r(\mathbf{W}^*)$

$$\|\widehat{\boldsymbol{\Omega}}(\mathbf{W}^*, \mathbf{W}) - \boldsymbol{\Omega}(\mathbf{W}^*, \mathbf{W})\|_2 \leq c_{10} K^{5/2} \sqrt{\frac{d \log N}{N}} \tag{B.7}$$

holds with probability at least $1 - d^{-10}$. $\square$

## B.3 Proof of Lemma A.4

*Proof.* Now we turn to bound the Frobenius norm of the error matrix $\mathbf{E}(\mathbf{W})$, which follows the similar argument of Lemma A.3. Consider the $j$-th column of $\mathbf{E}(\mathbf{W})$, we have

$$\mathbb{P}\bigg(\sup_{\mathbf{W}\in\mathcal{B}_r(\mathbf{W}^*)}\|\mathbf{E}_j(\mathbf{W})\|_2 \geq t\bigg) \leq \mathbb{P}\bigg(\sup_{\mathbf{W}\in\mathcal{B}_r(\mathbf{W}^*)}\|\mathbf{E}_j(\mathbf{W}) - \mathbf{E}_j(\widetilde{\mathbf{W}})\|_2 \geq \frac{t}{2}\bigg)$$
$$+ \mathbb{P}\bigg(\sup_{\widetilde{\mathbf{W}}\in\mathcal{N}_\zeta}\|\mathbf{E}_j(\widetilde{\mathbf{W}})\|_2 \geq \frac{t}{2}\bigg).$$

**Bounding $\sup_{\widetilde{\mathbf{W}}\in\mathcal{N}_\zeta}\|\mathbf{E}_j(\widetilde{\mathbf{W}})\|_2$:** We have

$$\mathbf{E}_j(\widetilde{\mathbf{W}}) = \frac{1}{N}\sum_{i=1}^N \epsilon_i \mathbf{x}_i \cdot \mathbb{1}\{\widetilde{\mathbf{w}}_j^\top \mathbf{x}_i\} := \frac{1}{N}\sum_{i=1}^N \epsilon_i \cdot \widehat{\mathbf{x}}_i,$$

where we denote $\widehat{\mathbf{x}}_i = \mathbf{x}_i \cdot \mathbb{1}\{\widetilde{\mathbf{w}}_j^\top \mathbf{x}_i\}$ for simplicity. Similarly, we can show that $\widehat{\mathbf{x}}_i$'s follow i.i.d. sub-Gaussian distribution with sub-Gaussian norm $C$. According to the definition of $\ell_2$-norm, we have

$$\|\mathbf{E}_j(\widetilde{\mathbf{W}})\|_2 = \sup_{\|\mathbf{u}\|_2=1}\bigg\langle \frac{1}{N}\sum_{i=1}^N \epsilon_i \widehat{\mathbf{x}}_i, \mathbf{u}\bigg\rangle = \sup_{\|\mathbf{u}\|_2=1}\frac{1}{N}\sum_{i=1}^N \mathbf{u}^\top \widehat{\mathbf{x}}_i \epsilon_i = \sup_{\|\mathbf{u}\|_2=1}\frac{1}{N}\mathbf{u}^\top \widehat{\mathbf{X}}\boldsymbol{\epsilon}, \quad \text{(B.8)}$$

where we denote $\widehat{\mathbf{X}} = [\widehat{\mathbf{x}}_1, \widehat{\mathbf{x}}_2, \ldots, \widehat{\mathbf{x}}_N] \in \mathbb{R}^{d\times N}$. In the following discussions, we are going to use the covering number argument and Bernstein-type inequality to bound the term $\|\mathbf{E}_j(\widetilde{\mathbf{W}})\|_2$. Let $\mathcal{N}_{\zeta'}$ be a $\zeta'$-net of unit Euclidean sphere $S^{d-1}$ such that for any $\mathbf{u} \in S^{d-1}$, there exists $\mathbf{v} \in \mathcal{N}_{\zeta'}$ satisfying $\|\mathbf{u}-\mathbf{v}\|_2 \leq \zeta'$, where $\zeta' \in (0,1)$ will be specified later. Choose $\mathbf{u}_0 \in S^{d-1}$ such that $\mathbf{u}_0^\top \widehat{\mathbf{X}}\boldsymbol{\epsilon}/N = \|\widehat{\mathbf{X}}\boldsymbol{\epsilon}/N\|_2$, then there exists $\mathbf{v}_0 \in \mathcal{N}_{\zeta'}$ satisfying $\|\mathbf{u}_0 - \mathbf{v}_0\|_2 \leq \zeta'$. Thus, we have

$$\frac{1}{N}\mathbf{v}_0^\top \widehat{\mathbf{X}}\boldsymbol{\epsilon} = \frac{1}{N}\mathbf{u}_0^\top \widehat{\mathbf{X}}\boldsymbol{\epsilon} - \frac{1}{N}(\mathbf{u}_0 - \mathbf{v}_0)^\top \widehat{\mathbf{X}}\boldsymbol{\epsilon} \geq (1-\zeta') \cdot \bigg\|\frac{1}{N}\widehat{\mathbf{X}}\boldsymbol{\epsilon}\bigg\|_2, \quad \text{(B.9)}$$

where the inequality follows from the Cauchy-Schwarz inequality. By the union bound, we have

$$\mathbb{P}\bigg(\sup_{\widetilde{\mathbf{W}}\in\mathcal{N}_\zeta}\|\mathbf{E}_j(\widetilde{\mathbf{W}})\|_2 \geq \frac{t}{2}\bigg) \leq |\mathcal{N}_\zeta|\mathbb{P}\bigg(\|\mathbf{E}_j(\widetilde{\mathbf{W}})\|_2 \geq \frac{t}{2}\bigg). \quad \text{(B.10)}$$

Furthermore, according to (B.8), we have

$$\mathbb{P}\bigg(\|\mathbf{E}_j(\widetilde{\mathbf{W}})\|_2 \geq \frac{t}{2}\bigg) \leq \mathbb{P}\bigg(\frac{1}{1-\zeta'} \cdot \sup_{\mathbf{u}\in\mathcal{N}_{\zeta'}}\frac{1}{N}\sum_{i=1}^N \epsilon_i \mathbf{u}^\top \widehat{\mathbf{x}}_i \geq \frac{t}{2}\bigg)$$
$$\leq \bigg(1+\frac{2}{\zeta'}\bigg)^d \cdot \mathbb{P}\bigg(\frac{1}{1-\zeta'} \cdot \frac{1}{N}\sum_{i=1}^N \epsilon_i \mathbf{u}^\top \widehat{\mathbf{x}}_i \geq \frac{t}{2}\bigg)$$
$$= 5^d \cdot \mathbb{P}\bigg(\frac{1}{N}\sum_{i=1}^N \epsilon_i \mathbf{u}^\top \widehat{\mathbf{x}}_i \geq \frac{t}{4}\bigg),$$

where we set $\zeta' = 1/2$ in the last equality, and the second inequality follows from the union bound and Lemma C.4. Since $\epsilon_i$ follows a centered sub-Gaussian distribution with sub-Gaussian norm $\nu$ and is independent with $\mathbf{x}_i$, we can obtain that $\epsilon_i$ is independent with $\mathbf{u}^\top \widehat{\mathbf{x}}_i$. In addition, we have derived that $\mathbf{u}^\top \widehat{\mathbf{x}}_i$ is a sub-Gaussian random variable with $\|\mathbf{u}^\top \widehat{\mathbf{x}}_i\|_{\psi_2} \leq C$. Therefore, we have $\{\epsilon_i \mathbf{u}^\top \widehat{\mathbf{x}}_i\}_{i=1}^N$ are i.i.d. sub-exponential random variables (See Appendix C for the formal definition of sub-exponential random variable) with sub-exponential norm $\|\epsilon_i \mathbf{u}^\top \widehat{\mathbf{x}}_i\|_{\psi_1} \leq 2\|\epsilon_i\|_{\psi_2} \cdot \|\mathbf{u}^\top \widehat{\mathbf{x}}_i\|_{\psi_2} \leq 2\nu C$. Furthermore, we have $\mathbb{E}[\epsilon_i \mathbf{u}^\top \widehat{\mathbf{x}}_i] = \mathbb{E}[\epsilon_i] \cdot \mathbb{E}[\mathbf{u}^\top \widehat{\mathbf{x}}_i] = 0$. Thus by Bernstein-type inequality for sub-exponential random variables in Lemma C.6, we have

$$\mathbb{P}\bigg(\frac{1}{N}\sum_{i=1}^N \epsilon_i \mathbf{u}^\top \widehat{\mathbf{x}}_i \geq \frac{t}{4}\bigg) \leq \exp\bigg[-c\min\bigg(\frac{Nt^2}{4(\nu C)^2}, \frac{Nt}{2\nu C}\bigg)\bigg].$$

Therefore, we can obtain

$$\mathbb{P}\bigg(\sup_{\|\mathbf{u}\|_2 = 1} \frac{1}{N}\mathbf{u}^\top \widehat{\mathbf{X}}\boldsymbol{\epsilon} \geq \frac{t}{2}\bigg) \leq \exp\bigg[-c_1\min\bigg(\frac{Nt^2}{4(\nu C)^2}, \frac{Nt}{2\nu C}\bigg) + d\log 5\bigg],$$

which implies that

$$\mathbb{P}\bigg(\|\mathbf{E}_j(\widetilde{\mathbf{W}})\|_2 \geq \frac{t}{2}\bigg) \leq \exp\bigg[-c_1\min\bigg(\frac{Nt^2}{4(\nu C)^2}, \frac{Nt}{2\nu C}\bigg) + d\log 5\bigg].$$

As a result, according to (B.10), we can obtain

$$\mathbb{P}\bigg(\sup_{\widetilde{\mathbf{W}} \in \mathcal{N}_\zeta} \|\mathbf{E}_j(\widetilde{\mathbf{W}})\|_2 \geq \frac{t}{2}\bigg) \leq \exp\big(dK\log(3r/\zeta) - c_1 Nt^2/(4\nu^2 C^2) + d\log 5\big),$$

where the inequality is due to Lemma C.5 such that $\log |\mathcal{N}_\zeta| \leq dK\log(3r/\zeta)$.

As long as $t \geq c_2 \nu \sqrt{\big(dK\log(3r/\zeta) + d\log 5 + \log(4K/\delta)\big)/N}$, we have

$$\mathbb{P}\bigg(\sup_{\widetilde{\mathbf{W}} \in \mathcal{N}_\zeta} \|\mathbf{E}_j(\widetilde{\mathbf{W}})\|_2 \geq \frac{t}{2}\bigg) \leq \frac{\delta}{2K}. \tag{B.11}$$

**Bounding $\sup_{\mathbf{W} \in \mathcal{B}_r(\mathbf{W}^*)} \|\mathbf{E}_j(\mathbf{W}) - \mathbf{E}_j(\widetilde{\mathbf{W}})\|_2$:** By the definition of $\mathbf{E}(\mathbf{W})$, we have

$$\mathbf{E}_j(\mathbf{W}) - \mathbf{E}_j(\widetilde{\mathbf{W}}) = \frac{1}{N}\sum_{i=1}^N \epsilon_i \mathbf{x}_i \cdot \big(\mathbb{1}\{\mathbf{w}_j^\top \mathbf{x}_i\} - \mathbb{1}\{\widetilde{\mathbf{w}}_j^\top \mathbf{x}_i\}\big),$$

which implies

$$\sup_{\mathbf{W} \in \mathcal{B}_r(\mathbf{W}^*)} \|\mathbf{E}_j(\mathbf{W}) - \mathbf{E}_j(\widetilde{\mathbf{W}})\|_2 \leq \max_i \|\epsilon_i \mathbf{x}_i\|_2 \cdot \sup_{\mathbf{W} \in \mathcal{B}_r(\mathbf{W}^*)} \mathbb{1}\{\mathbf{w}_j^\top \mathbf{x}_i \geq 0\}\mathbb{1}\{\widetilde{\mathbf{w}}_j^\top \mathbf{x}_i \leq 0\}$$

$$+ \max_i \|\epsilon_i \mathbf{x}_i\|_2 \cdot \sup_{\mathbf{W} \in \mathcal{B}_r(\mathbf{W}^*)} \mathbb{1}\{\widetilde{\mathbf{w}}_j^\top \mathbf{x}_i \geq 0\}\mathbb{1}\{\mathbf{w}_j^\top \mathbf{x}_i \leq 0\}.$$

Thus according to (B.5), we have

$$\mathbb{P}\bigg(\sup_{\mathbf{W} \in \mathcal{B}_r(\mathbf{W}^*)} \mathbb{1}\{\mathbf{w}_j^\top \mathbf{x}_i \geq 0\}\mathbb{1}\{\widetilde{\mathbf{w}}_j^\top \mathbf{x}_i \leq 0\} \geq t_1\bigg) \leq c_3\frac{\zeta}{t_1}.$$

In addition, we have

$$\mathbb{P}\bigg(\max_i \|\epsilon_i \mathbf{x}_i\|_2 \geq t_2\bigg) \leq 2N \exp(-c_4 t_2^2/\nu^2).$$

Hence we can obtain

$$\mathbb{P}\bigg(\max_i \|\epsilon_i \mathbf{x}_i\|_2 \cdot \sup_{\mathbf{W} \in \mathcal{B}_r(\mathbf{W}^*)} \mathbb{1}\{\mathbf{w}_j^\top \mathbf{x}_i \geq 0\} \mathbb{1}\{\widetilde{\mathbf{w}}_j^\top \mathbf{x}_i \leq 0\} \geq t_1 t_2 \bigg) \leq 2N \exp(-c_4 t_2^2/\nu^2) + c_3 \frac{\zeta}{t_1}.$$

As a result, we have

$$\mathbb{P}\bigg(\sup_{\mathbf{W} \in \mathcal{B}_r(\mathbf{W}^*)} \|\mathbf{E}_j(\mathbf{W}) - \mathbf{E}_j(\widetilde{\mathbf{W}})\|_2 \geq 2 t_1 t_2 \bigg) \leq 4N \exp(-c_4 t_2^2/\nu^2) + 2 c_3 \frac{\zeta}{t_1}.$$

Therefore, let $t_1 \geq c_5 \nu \sqrt{\log(NK/\delta)}$, $t_2 \geq c_6 \zeta K/\delta$, and replace $2 t_1 t_2$ with $t/2$, we can obtain that as long as $t \geq c_7 \nu \zeta K \sqrt{\log(NK/\delta)}/\delta$, the following holds

$$\mathbb{P}\bigg(\sup_{\mathbf{W} \in \mathcal{B}_r(\mathbf{W}^*)} \|\mathbf{E}_j(\mathbf{W}) - \mathbf{E}_j(\widetilde{\mathbf{W}})\|_2 \geq \frac{t}{2} \bigg) \leq \frac{\delta}{2K}. \quad (B.12)$$

Combining (B.11) and (B.12), we have

$$\mathbb{P}\bigg(\sup_{\mathbf{W} \in \mathcal{B}_r(\mathbf{W}^*)} \|\mathbf{E}_j(\mathbf{W})\|_2 \geq t \bigg) \leq \frac{\delta}{K}.$$

Since we have

$$\|\mathbf{E}(\mathbf{W})\|_F \leq \bigg(\sum_{j=1}^K \|\mathbf{E}_j(\mathbf{W})\|_2^2\bigg)^{1/2},$$

we can obtain

$$\mathbb{P}\bigg(\sup_{\mathbf{W} \in \mathcal{B}_r(\mathbf{W}^*)} \|\mathbf{E}(\mathbf{W})\|_F \geq \sqrt{K} t\bigg) \leq \delta.$$

Therefore, let $\delta = d^{-10}$, $\zeta = c_8 r \delta/N$, we have $t \geq c_9 \nu K \sqrt{d \log N/N}$, and we can obtain

$$\|\mathbf{E}(\mathbf{W})\|_F \leq \bigg(\sum_{j=1}^K \|\mathbf{E}_j(\mathbf{W})\|_2^2\bigg)^{1/2} \leq c_9 \nu K^{3/2} \sqrt{\frac{d \log N}{N}} \quad (B.13)$$

holds with probability at least $1 - d^{-10}$. □

## C Additional Definitions and Lemmas

For self-containedness of the paper, in this section, we provide additional definitions and lemmas used in the previous sections. For more detailed discussions, please refer to Vershynin (2010).

**Definition C.1.** (Vershynin, 2010) (sub-Gaussian random vector) We say $\boldsymbol{X} \in \mathbb{R}^d$ is a sub-Gaussian random vector, if the one-dimensional marginals $\boldsymbol{\alpha}^\top \boldsymbol{X}$ are sub-Gaussian random variables for all unit vector $\boldsymbol{\alpha} \in S^{d-1}$. The sub-Gaussian norm of $\boldsymbol{X}$ is defined as $\|\boldsymbol{X}\|_{\psi_2} = \sup_{\|\boldsymbol{\alpha}\|_2=1} \|\boldsymbol{\alpha}^\top \boldsymbol{X}\|_{\psi_2}$.

**Definition C.2.** (Vershynin, 2010)(sub-exponential random variable) We say $X$ is a sub-exponential random variable with sub-exponential norm $K > 0$, if $(\mathbb{E}|X|^p)^{1/p} \leq Kp$ for all $p \geq 1$. In addition, the sub-exponential norm of $X$, denoted $\|X\|_{\psi_1}$, is defined as $\|X\|_{\psi_1} = \sup_{p \geq 1} p^{-1}(\mathbb{E}|X|^p)^{1/p}$.

**Lemma C.3.** (Vershynin, 2010) Let $\mathbf{A}$ be an $N \times n$ whose rows $\mathbf{A}_i$ are independent sub-Gaussian random vectors in $\mathbb{R}^n$ with common second moment matrix $\boldsymbol{\Sigma}$. Then for every $t \geq 0$, the following inequality holds with probability at least $1 - 2\exp(-ct^2)$:

$$\left\|\frac{1}{N}\mathbf{A}^\top \mathbf{A} - \boldsymbol{\Sigma}\right\|_2 \leq \max(\zeta, \zeta^2), \quad \text{where } \zeta = C\sqrt{\frac{n}{N}} + \frac{t}{\sqrt{N}}.$$

Here $C = C_K, c = c_K > 0$ depend only on the subgaussian norm $K = \max_i \|\mathbf{A}_i\|_{\psi_2}$ of the rows.

**Lemma C.4.** (Vershynin, 2010) The unit Euclidean sphere $S^{n-1}$ equipped with the Euclidean metric satisfies for every $\epsilon > 0$ that

$$\mathcal{N}(S^{n-1}, \epsilon) \leq \left(1 + \frac{2}{\epsilon}\right)^n.$$

**Lemma C.5.** For the unit norm ball $\mathcal{B}$ in $\mathbb{R}^d$ equipped with the Euclidean metric, we have for every $\epsilon > 0$ that

$$\mathcal{N}(\mathcal{B}, \epsilon) \leq \left(1 + \frac{2}{\epsilon}\right)^d.$$

**Lemma C.6.** (Vershynin, 2010) Let $X_1, \ldots, X_N$ be independent centered sub-exponential random variables, and $K = \max_i \|X_i\|_{\psi_1}$. Then for every $\mathbf{a} = (a_1, \ldots, a_N) \in \mathbb{R}^N$ and every $t \geq 0$, we have

$$\mathbb{P}\bigg\{\sum_{i=1}^N a_i X_i \geq t\bigg\} \leq \exp\bigg[-c\min\bigg(\frac{t^2}{K^2\|\mathbf{a}\|_2^2}, \frac{t}{K\|\mathbf{a}\|_\infty}\bigg)\bigg],$$

where $c > 0$ is an absolute constant.

**Lemma C.7.** (Du et al., 2017a) If $\boldsymbol{X} \sim N(0, \mathbf{I})$, for $\theta \in [0, \pi/2]$, we have

$$\max_{\angle \mathbf{w}, \widetilde{\mathbf{w}} = \theta} \lambda_{\max}\big(\boldsymbol{\Sigma}(\mathbf{w}, -\widetilde{\mathbf{w}})\big) \leq d\theta,$$

where $\boldsymbol{\Sigma}(\mathbf{w}, -\widetilde{\mathbf{w}}) = \mathbb{E}_{\boldsymbol{X} \sim \mathcal{D}_X}[\widehat{\boldsymbol{\Sigma}}_{\mathbf{w}, -\widetilde{\mathbf{w}}}]$, and $\widehat{\boldsymbol{\Sigma}}_{\mathbf{w}, -\widetilde{\mathbf{w}}}$ is defined in (3.3).